\documentclass[letterpaper]{article} % DO NOT CHANGE THIS
\usepackage{aaai2026}  % DO NOT CHANGE THIS [submission]
\usepackage{times}  % DO NOT CHANGE THIS
\usepackage{helvet}  % DO NOT CHANGE THIS
\usepackage{courier}  % DO NOT CHANGE THIS
\usepackage[hyphens]{url}  % DO NOT CHANGE THIS
\usepackage{graphicx} % DO NOT CHANGE THIS
\usepackage{natbib}  % DO NOT CHANGE THIS AND DO NOT ADD ANY OPTIONS TO IT
\usepackage{caption} % DO NOT CHANGE THIS AND DO NOT ADD ANY OPTIONS TO IT
\usepackage{algorithm}
\usepackage{algorithmic}

\usepackage{latexsym}
\usepackage{booktabs}
\usepackage{makecell}
\usepackage{multirow}
\usepackage{graphicx}
\usepackage{todonotes}
\usepackage{listings}
\usepackage{subcaption}

\usepackage{colortbl}

\usepackage{amsmath}
\usepackage{amsfonts}
\usepackage{amssymb}
\usepackage{bm}

\usepackage{newfloat}
\usepackage{listings}
\DeclareCaptionStyle{ruled}{labelfont=normalfont,labelsep=colon,strut=off} % DO NOT CHANGE THIS
\floatstyle{ruled}
\newfloat{listing}{tb}{lst}{}
\floatname{listing}{Listing}
\nocopyright %-- Your paper will not be published if you use this command
\title{Sparse-dLLM: Accelerating Diffusion LLMs with Dynamic Cache Eviction}
\author{
    Yuerong Song\textsuperscript{1,2},
    Xiaoran Liu\textsuperscript{1,2},
    Ruixiao Li\textsuperscript{1,2},
    Zhigeng Liu\textsuperscript{1,2}, \\
    Zengfeng Huang\textsuperscript{1,2},
    Qipeng Guo\textsuperscript{2,3},
    Ziwei He\textsuperscript{2}\thanks{\ \ Corresponding Author. },
    Xipeng Qiu\textsuperscript{1,2}\footnotemark[1]
}
\affiliations{
    \textsuperscript{1}School of Computer Science, Fudan University,
    \textsuperscript{2}Shanghai Innovation Institute, 
    \textsuperscript{3}Shanghai AI Lab \\
    \texttt{yuerongsong25@m.fudan.edu.cn}, 
    \texttt{xpqiu@fudan.edu.cn}, \texttt{ziweihe@outlook.com}\\

}

\usepackage{bibentry}
\begin{document}

\maketitle

\begin{abstract}
Diffusion Large Language Models (dLLMs) enable breakthroughs in reasoning and parallel decoding but suffer from prohibitive quadratic computational complexity and memory overhead during inference. Current caching techniques accelerate decoding by storing full-layer states, yet impose substantial memory usage that limit long-context applications. Our analysis of attention patterns in dLLMs reveals persistent cross-layer sparsity, with pivotal tokens remaining salient across decoding steps and low-relevance tokens staying unimportant, motivating selective cache eviction. We propose Sparse-dLLM, the first training-free framework integrating dynamic cache eviction with sparse attention via delayed bidirectional sparse caching. By leveraging the stability of token saliency over steps, it retains critical tokens and dynamically evicts unimportant prefix/suffix entries using an attention-guided strategy. Extensive experiments on LLaDA and Dream series demonstrate Sparse-dLLM achieves up to 10$\times$ higher throughput than vanilla dLLMs, with comparable performance and similar peak memory costs, outperforming previous methods in efficiency and effectiveness. The code is available
at \url{https://github.com/OpenMOSS/Sparse-dLLM}.

\end{abstract}

\section{Introduction}

Diffusion Large Language Models, or dLLMs, have garnered significant attention in the Natural Language Processing community~\citep{nie2025large,dream2025}. They are seen as a promising approach to addressing key limitations of traditional auto-regressive LLMs~\citep{touvron2023llama,Sun2024MOSS}, such as the reversal curse~\citep{berglund2023reversal}, and enabling advanced reasoning~\citep{dziri2023faith,dream2025}, and parallel decoding~\citep{mercury,genimid}. Extensive research has been dedicated to exploring their scalability~\citep{nie2025large,dream2025}, adapting them for multi-modal applications~\citep{yang2025mmada,you2025llada}, and adapting them for reasoning tasks~\citep{zhao2025d1,huang2025reinforcing,zhu2025llada}. However, current open-source dLLMs show a significant throughput shortage in practice, with their actual speed lags behind that of auto-regressive LLMs~\citep{ma2025dkv,hu2025accelerating,wu2025fast,liudllm}.

\begin{figure}[!tb]
\centering
\includegraphics[width=0.75\linewidth]{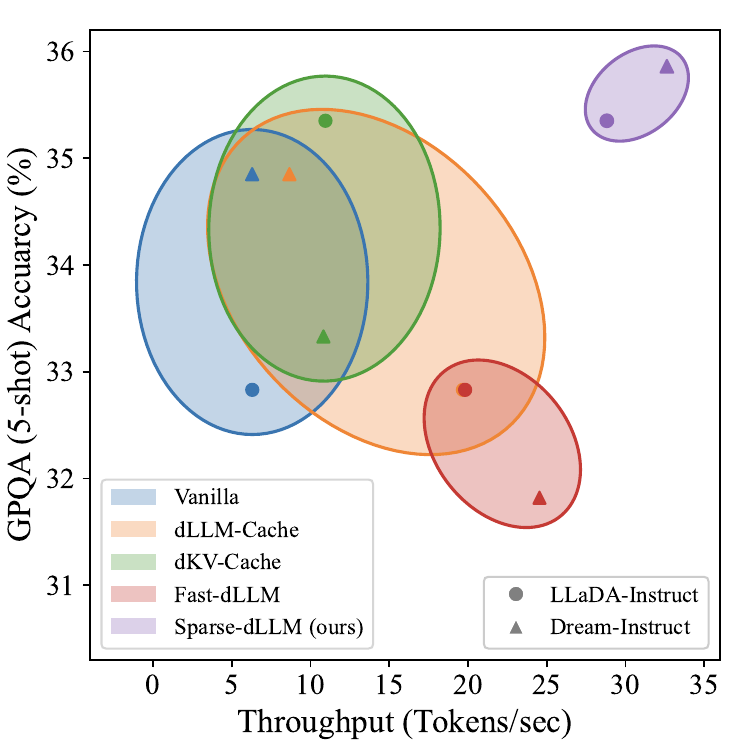}
\caption{Throughput vs. Accuracy across methods. Sparse-dLLM (ours) achieves the best throughput while maintaining or even improving performance of vanilla dLLMs.}
\label{fig_intro}
\end{figure}

\begin{figure*}[!tb]
\centering
\begin{subfigure}[b]{0.24\linewidth}
    \centering
    \includegraphics[width=0.92\linewidth]{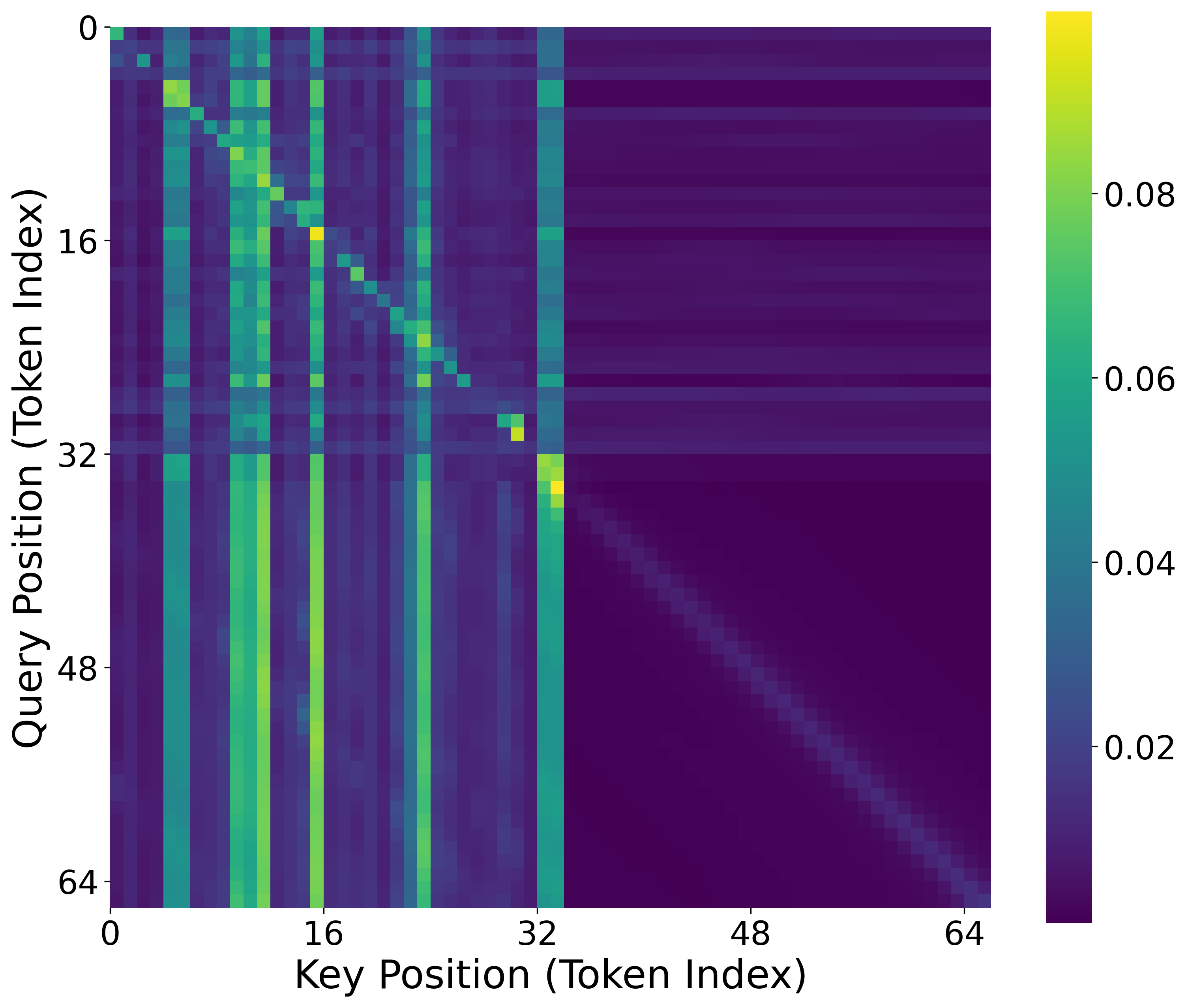}
    \caption{Layer 0, Step 0}
    \label{fig_sparsity_0_0}
\end{subfigure}
\hfill
\begin{subfigure}[b]{0.24\linewidth}
    \centering
    \includegraphics[width=0.92\linewidth]{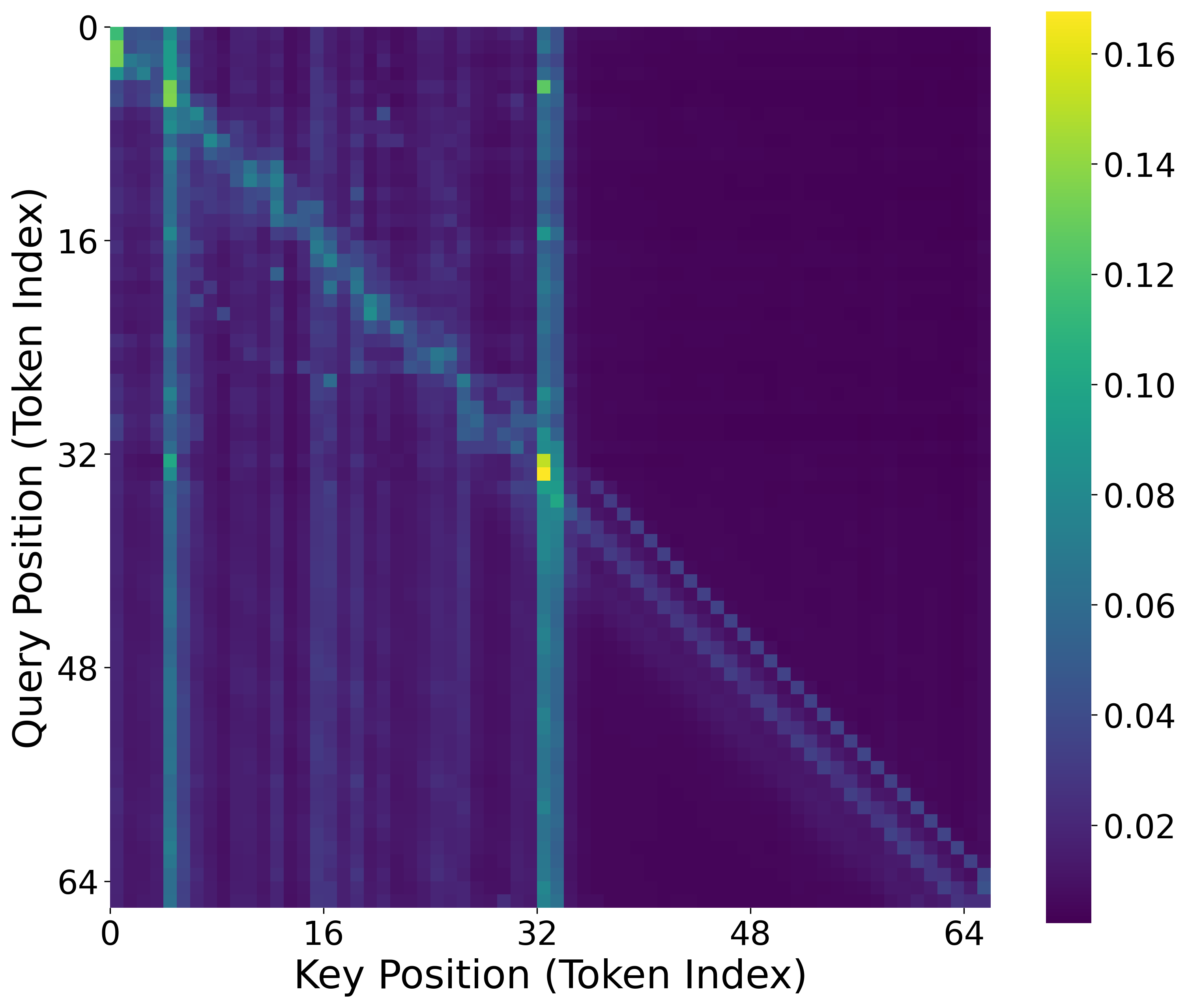}
    \caption{Layer 4, Step 0}
    \label{fig_sparsity_4_0}
\end{subfigure}
\hfill
\begin{subfigure}[b]{0.24\linewidth}
    \centering
    \includegraphics[width=0.92\linewidth]{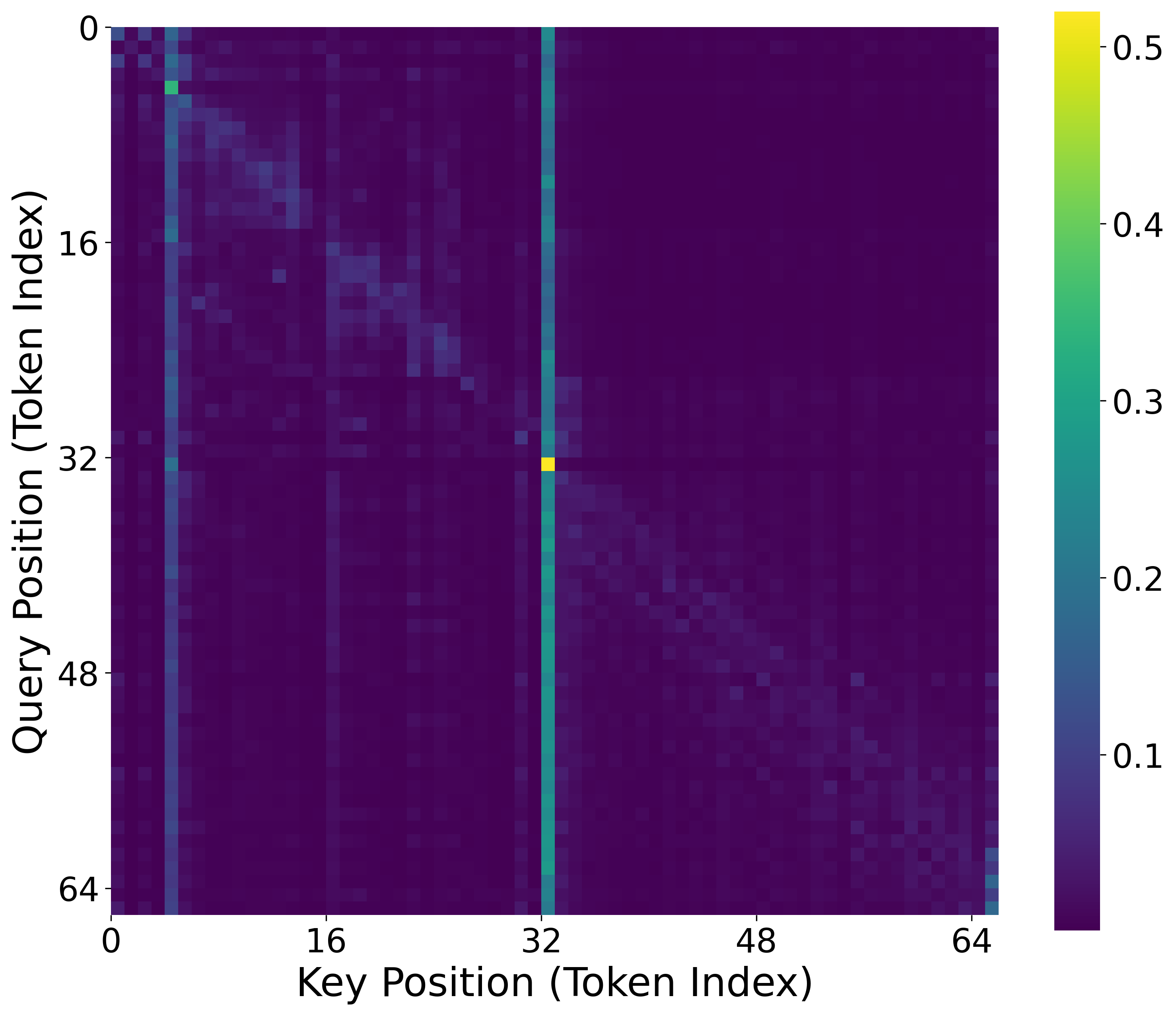}
    \caption{Layer 15, Step 0}
    \label{fig_sparsity_15_0}
\end{subfigure}
\hfill
\begin{subfigure}[b]{0.24\linewidth}
    \centering
    \includegraphics[width=0.92\linewidth]{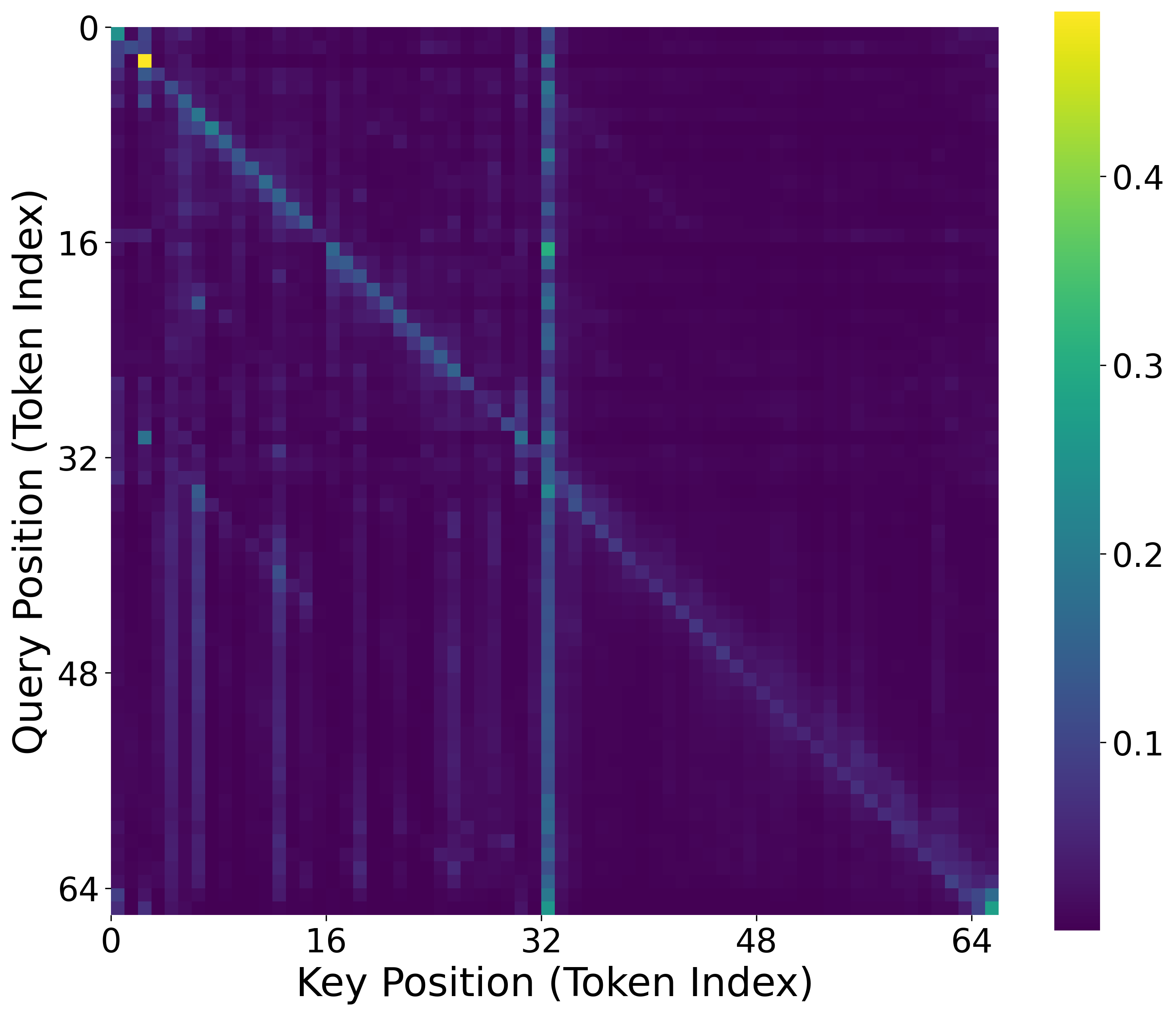}
    \caption{Layer 31, Step 0}
    \label{fig_sparsity_31_0}
\end{subfigure}
\\[0.6ex]
\begin{subfigure}[b]{0.24\linewidth}
    \centering
    \includegraphics[width=0.92\linewidth]{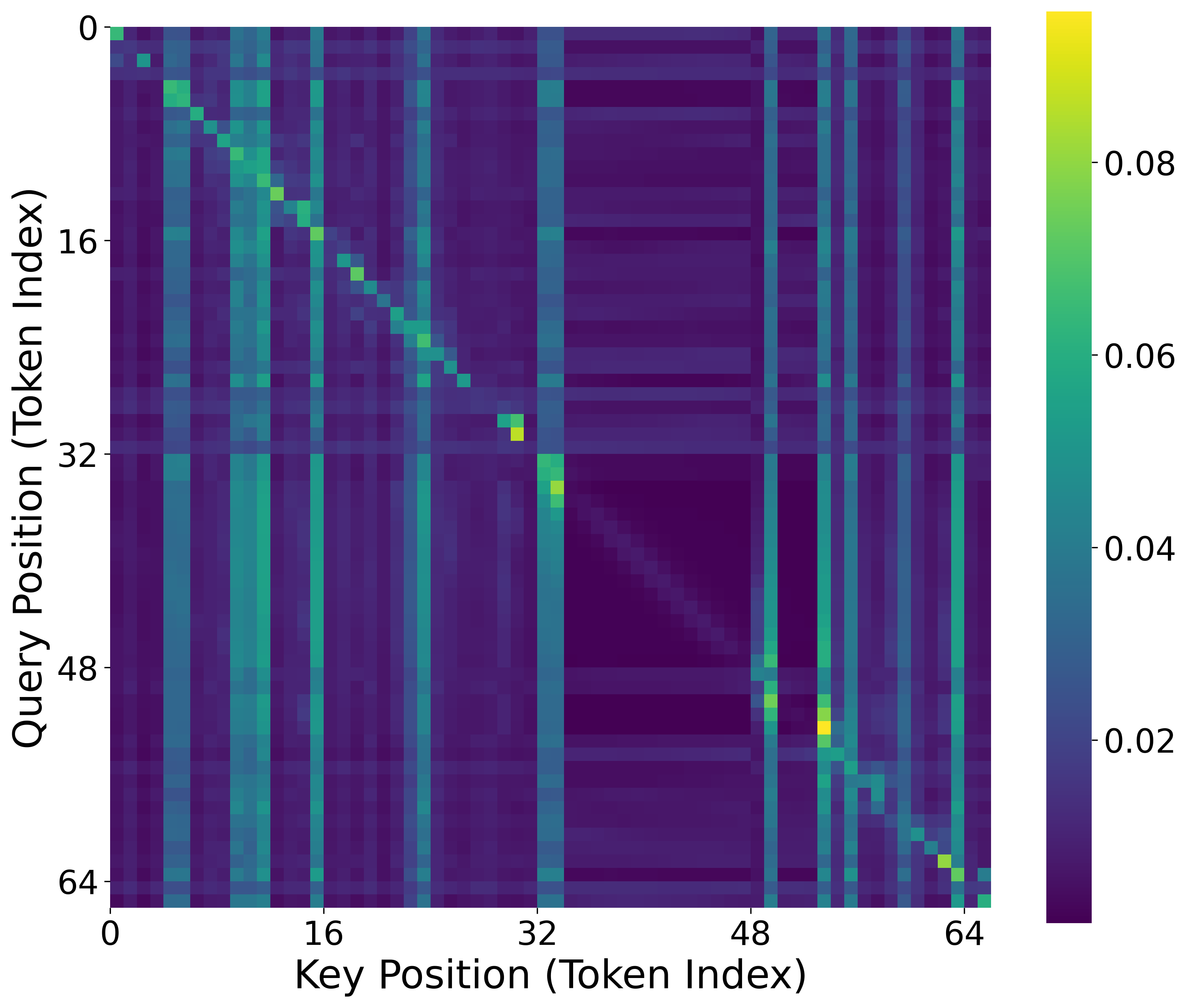}
    \caption{Layer 0, Step 15}
    \label{fig_sparsity_0_15}
\end{subfigure}
\hfill
\begin{subfigure}[b]{0.24\linewidth}
    \centering
    \includegraphics[width=0.92\linewidth]{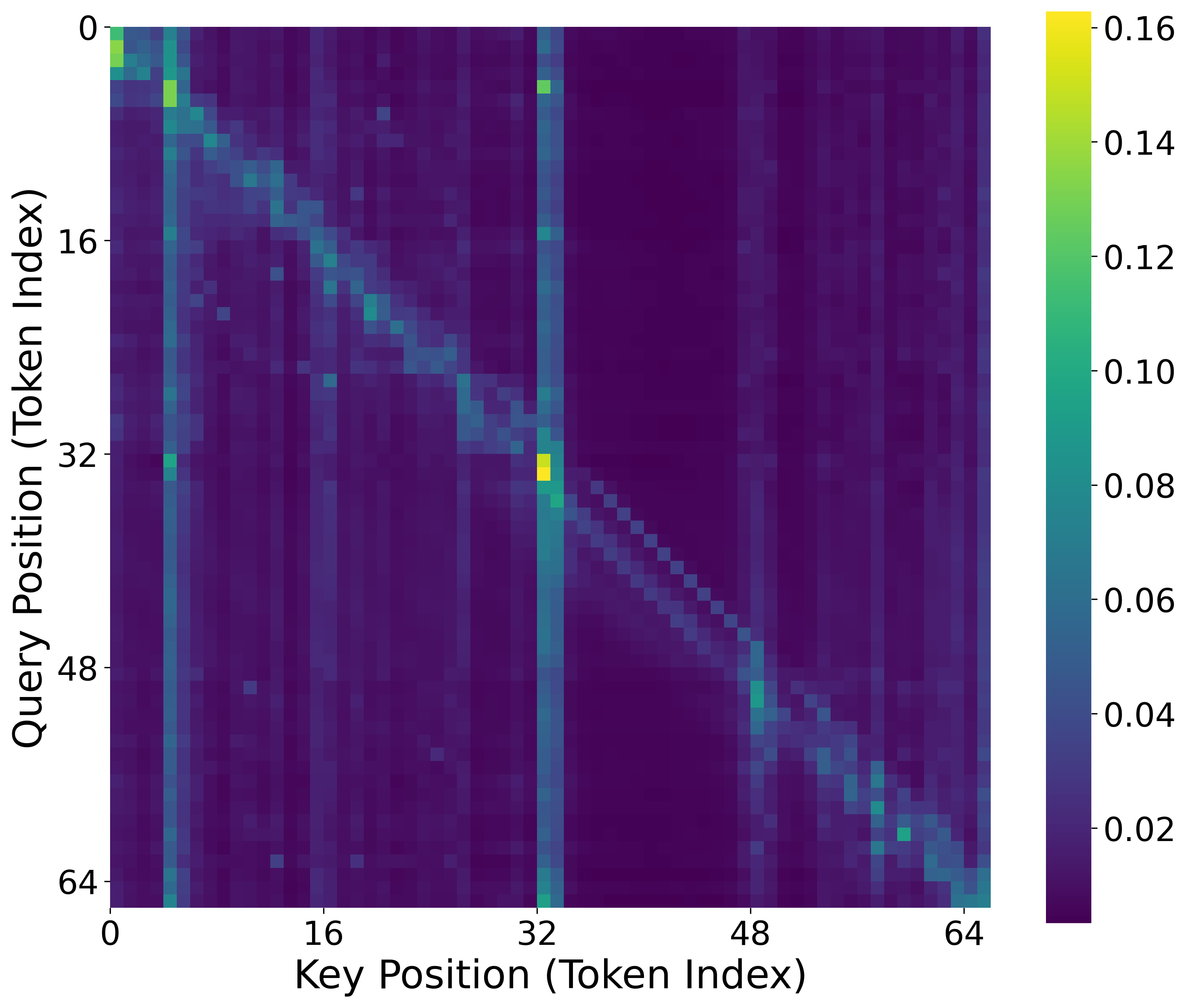}
    \caption{Layer 4, Step 15}
    \label{fig_sparsity_4_15}
\end{subfigure}
\hfill
\begin{subfigure}[b]{0.24\linewidth}
    \centering
    \includegraphics[width=0.92\linewidth]{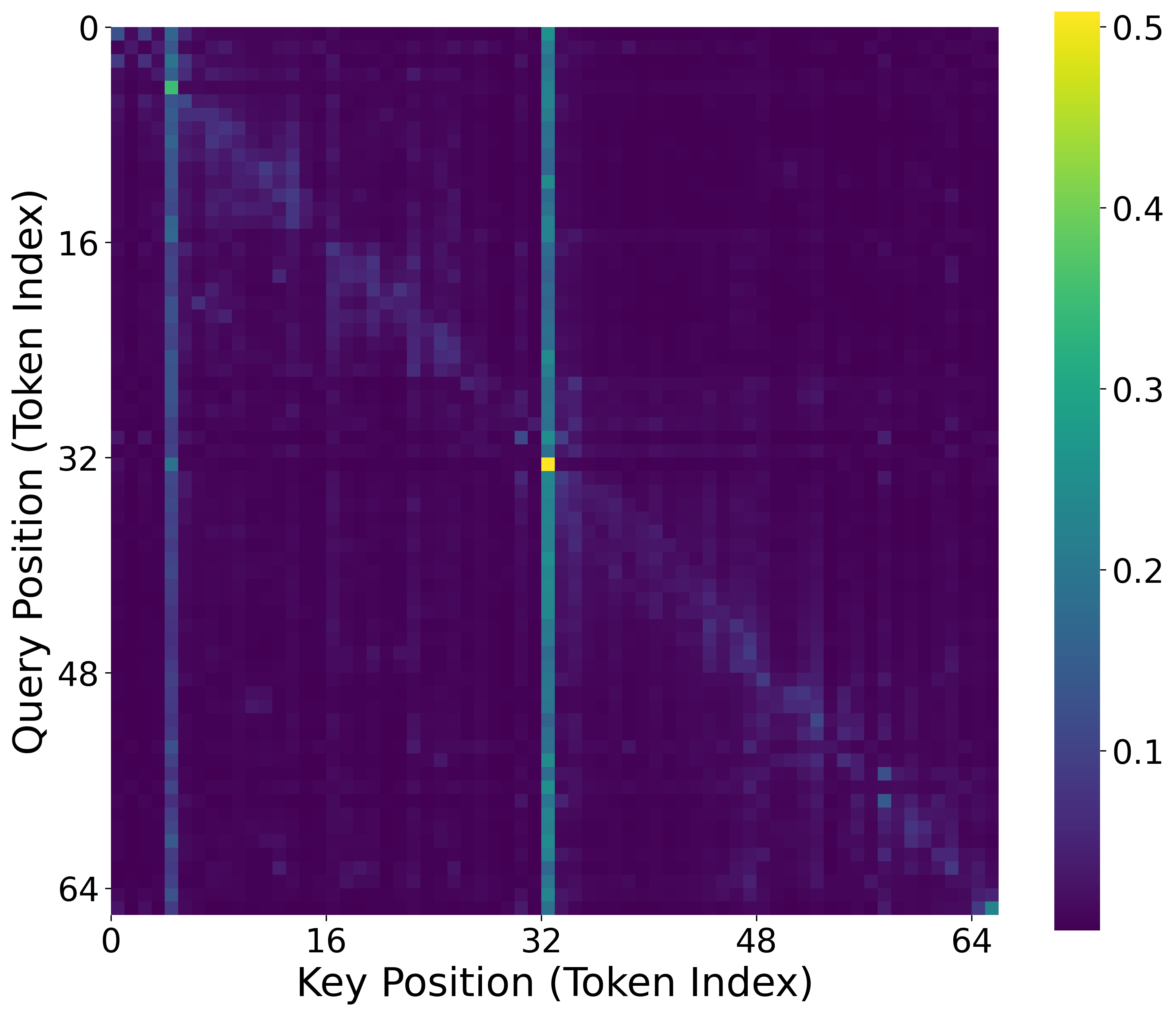}
    \caption{Layer 15, Step 15}
    \label{fig_sparsity_15_15}
\end{subfigure}
\hfill
\begin{subfigure}[b]{0.24\linewidth}
    \centering
    \includegraphics[width=0.92\linewidth]{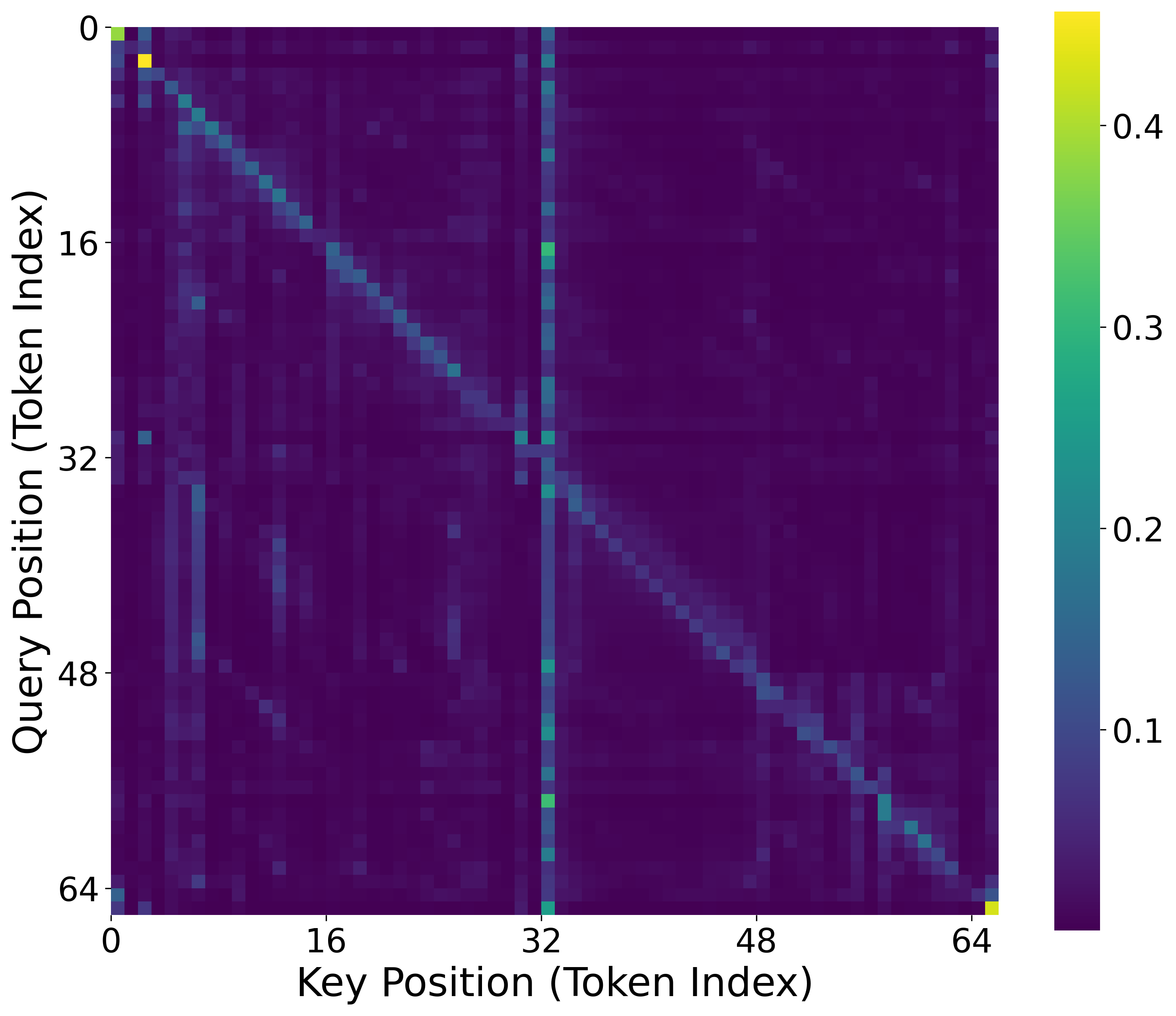}
    \caption{Layer 31, Step 15}
    \label{fig_sparsity_31_15}
\end{subfigure}
\caption{Sparsity patterns in dLLM attention. Using LLaDA-8B-Instruct with L = 66, T = 32, and a block length of 32, we observe pronounced sparsity that persists across layers, with pivotal tokens remaining salient throughout decoding steps.}
\label{fig_sparsity}
\end{figure*}

While traditional auto-regressive LLMs exhibit $\mathcal{O}(L)$ computational complexity during decoding (where $L$ denotes the prompt length), dLLMs incur a significantly higher $\mathcal{O}(L^2)$ complexity. This stems from the requirement to recompute the QKV states for the entire sequence, including the input prompt, all generated tokens, and mask tokens at every inference step. To address this cost, recent studies have adapted the KV cache mechanism from auto-regressive LLMs to dLLMs~\citep{ma2025dkv,wu2025fast,liudllm}. These approaches leverage the observation that the KV states across consecutive decoding steps are often nearly identical ~\citep{ma2025dkv,liudllm}. Consequently, they allow multiple steps to reuse the same cached KV states, accelerating decoding without compromising output quality. While this reuse strategy achieves computational savings by storing the complete KV representations for the sequence across all layers, this substantial memory overhead hinders the practical deployment of dLLMs for long-context scenarios.

Motivated by this limitation, we conduct a thorough analysis of attention patterns in dLLMs. As shown in Figure~\ref{fig_sparsity}, dLLMs exhibit significant sparsity, akin to auto-regressive LLMs, characterized by a sharp concentration on local positions and the vertical attention pattern~\citep{jiang2024minference} (where certain pivotal tokens remain salient across queries and decoding steps). Note that, unlike the causal attention of auto-regressive LLMs, dLLMs' bidirectional attention does not exhibit abnormal focus on initial tokens~\citep{xiao2023efficient}. Crucially, we find that although QKV states are recomputed at every inference step, the specific ones receiving significant attention remain remarkably stable across steps, indicating that low-salience tokens identified in early steps persistently exhibit minimal relevance throughout decoding. These observations—sparsity and stable key tokens—motivate our strategy: selectively evicting unimportant KV cache entries while retaining only the critical subset in early steps at each layer. Our approach markedly improves dLLM computational efficiency while introducing only minimal additional memory overhead, and without degrading downstream performance.

\begin{figure}[!tb]
\centering
\includegraphics[width=0.9\linewidth]{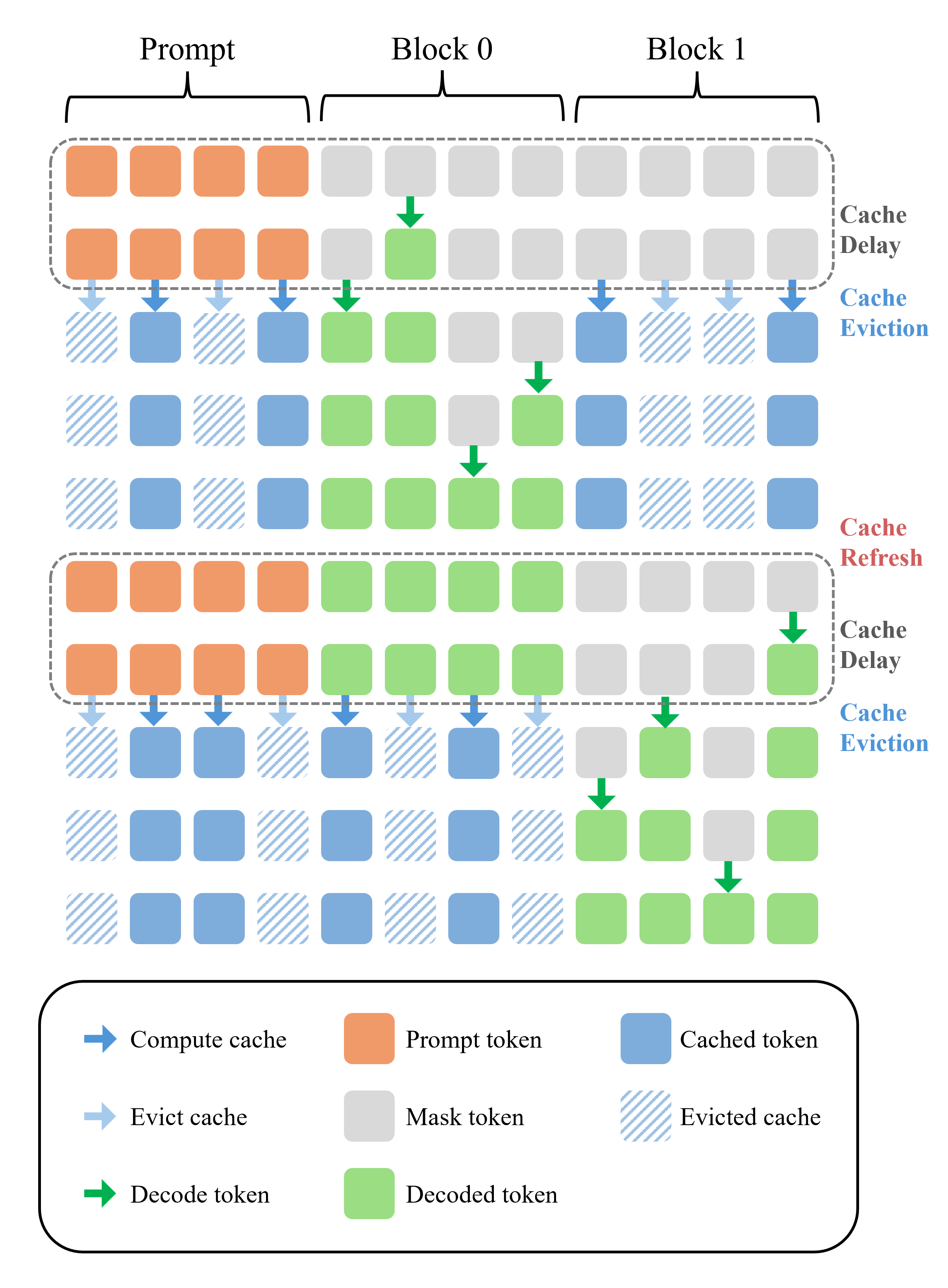}
\caption{Overview of Sparse-dLLM.}
\label{fig_main}
\end{figure}

We introduce \textbf{Sparse-dLLM}: the first training-free framework to integrate dynamic cache eviction with sparse attention for dLLMs. As illustrated in Figure~\ref{fig_main}, our approach employs a delayed bidirectional sparse strategy. First, we cache KV states for tokens outside the current decoding block during early inference step. Notably, we delay cache updates by one step to ensure stability. Then, leveraging temporal consistency in token saliency, we dynamically evict low-importance KV entries for both prefix and suffix tokens, guided by attention-aware sparse patterns with a pre-defined retention ratio. Cache states are fully refreshed when transitioning between blocks~\citep{wu2025fast}. As evidenced in Figure~\ref{fig_intro}, Sparse-dLLM achieves the best throughput while maintaining or even enhancing performance of vanilla dLLMs on certain tasks. Critically, by pruning redundant cache entries, our method maintains near-identical peak memory costs to vanilla dLLMs while optimizing computational efficiency. Our contributions are:

\begin{itemize}
\item We establish a formal analysis of sparsity patterns in dLLMs, revealing persistent cross-layer attention sparsity, with pivotal tokens remaining salient across decoding steps and low-relevance tokens staying unimportant, motivating our selective cache eviction strategy.
\item We propose Sparse-dLLM, the first training-free dynamic cache eviction method for dLLMs, featuring novel delayed bidirectional sparse caching that enables plug-and-play inference acceleration.
\item Extensive experiments on LLaDA and Dream demonstrate Sparse-dLLM achieves up to 10$\times$ higher throughput than vanilla dLLMs, while maintaining comparable performance and nearly identical memory costs, exceeding previous methods in efficiency and effectiveness. 
\end{itemize}

\section{Related Work}

KV cache optimization is critical for auto-regressive LLMs. Causal attention in AR maintains KV states for the input and generated tokens, allowing for direct caching and trading memory for computation. However, cache size grows with input length, limiting long-context deployment and driving KV cache optimization, whose typical strategy is token eviction. Current methods for KV cache sparsification in auto-regressive LLMs are retrospective, using fixed rules~\citep{xiao2023efficient}, past attention scores~\citep{zhang2024h2o,ge2023model}, or filtering based on a portion of previous tokens like SnapKV~\citep{li2025snapkv}, to manage already-generated tokens. Unlike auto-regressive LLMs, which can only see previously generated tokens, dLLMs can see the complete sequence. Therefore, our method designed for dLLMs retrospectively and prospectively sparsifies the cache, considering both prefix and suffix entries. These entries represent the cache states for the tokens preceding and succeeding the current block, respectively.

Although bidirectional attention in dLLMs prevents direct caching, recent studies leverage the observation that KV states across consecutive decoding steps are often nearly identical. Consequently, they adapt the KV cache mechanism from auto-regressive LLMs to dLLMs, which accelerates decoding without compromising output quality, as in dLLM-Cache~\citep{liudllm}, dKV-Cache~\citep{ma2025dkv}, FreeCache~\citep{hu2025accelerating}, and Fast-dLLM~\citep{wu2025fast}. Specifically, dLLM-Cache~\citep{liudllm} sets different refresh intervals for prompt cache and response cache, and uses feature similarity to update response partially. dKV-Cache~\citep{ma2025dkv} implements one-step delayed caching, where decoded tokens are cached not at their current decoding step but at the subsequent step, combined with a refreshing mechanism. Based on the rapidly diminishing contribution from masked tokens to earlier unmasked tokens, FreeCache~\citep{hu2025accelerating} caches prompt tokens' KV states. Fast-dLLM~\citep{wu2025fast} caches all KV states excluding the current decoding block. However, these approaches merely introduce a KV cache in dLLMs without examining its internal properties or further sparsifying it.

\section{Method}

\subsection{Preliminary: Inference of dLLM}

Unlike auto-regressive LLMs, diffusion language models (dLLMs) employ an iterative unmasking process to generate text through $T$ discrete decoding steps, progressively transforming a fully masked initial sequence into the final output. Using LLaDA~\citep{nie2025large} as an example, we formalize this process as follows.

Let $\mathbf{\mathcal{V}}$ be the vocabulary, $[\text{MASK}]\in\mathbf{\mathcal{V}}$ be the special mask token, $\mathbf{x}^t \in \mathbf{\mathcal{V}}^L$ be the sequence state at step $t$ for $t = T, \cdots, 0$. The initial state is defined as:
\begin{equation*}
    \mathbf{x}^T = (\mathbf{c}_0,\cdots,\mathbf{c}_{p-1},[\text{MASK}],\cdots,[\text{MASK}])
\end{equation*}
where $(\mathbf{c}_0,\cdots,\mathbf{c}_{p-1})$ constitute the prompt and $L$ denotes the total sequence length with $L-p$ mask tokens. 

At each step $t = T,\cdots,1$, a mask predictor model $f_{\theta}$ computes logits for the entire sequence:
\begin{equation*}
\mathbf{z}^t = f_{\theta}(\mathbf{x}^t)
\end{equation*}
Then, greedy decoding is performed on $\mathbf{z}^t$ to derive the predicted tokens $\hat{\mathbf{x}}^t$ for all masked positions:
\begin{equation*}
\hat{\mathbf{x}}_i^t = \underset{v \in \mathbf{\mathcal{V}}}{\arg\max}\ (\text{Softmax}(\mathbf{z}_i^t))_v \quad \text{if}\  \mathbf{x}_i^t == [\text{MASK}]
\end{equation*}
Finally, The transition function $S$~\citep{liudllm} selectively updates tokens in $\mathbf{x}^t$ based on predicted tokens $\hat{\mathbf{x}}^t$ (e.g., random or by confidence) to generate $\mathbf{x}^{t-1}$:
\begin{equation*}
\mathbf{x}^{t-1} = S(\hat{\mathbf{x}}^t, \mathbf{x}^t)
\end{equation*}

After $T$ steps, the final generated sequence $\mathbf{x}^0$ contains no mask tokens. However, iterative recomputation of all attention states for the full sequence imposes a substantial computational overhead, markedly increasing inference latency.

\subsection{Observations}

To unlock the potential of KV cache optimization for memory-efficient dLLM inference acceleration, we begin by systematically analyzing the attention patterns. As shown in Figure~\ref{fig_sparsity}, our analysis reveals two fundamental features: 

\paragraph{Sparsity Across Layers}
Horizontally in Figure~\ref{fig_sparsity}, we observe consistent sparsity across all dLLM layers within individual steps. Unlike auto-regressive models, dLLMs show no abnormal initial-token focus, but exhibit two stable patterns: (1) Local attention (bright diagonals) with strong neighbor focus, and (2) Vertical attention (bright verticals) where all queries concentrate on few pivotal keys. These patterns persist uniformly across layers, with most positions receiving minimal weights. 

\paragraph{Consistency Across Steps}

Vertically in Figure~\ref{fig_sparsity}, attention patterns across inference steps reveal remarkable temporal consistency in token saliency for tokens outside the current block. Take attention maps in Layer 0 as an example, as shown in Figures~\ref{fig_sparsity_0_0} and \ref{fig_sparsity_0_15}. Although QKV states are recomputed at every inference step, the specific tokens receiving significant attention remain remarkably stable across steps. This suggests that low-salience tokens outside the current block, once identified, consistently show minimal relevance throughout the decoding process.

These observations motivate us to selectively retain only the critical entries in the KV cache while evicting unimportant ones. Notably, while the local attention pattern is stable, we do not specifically use it for KV cache optimization, as its effective window in dLLMs is much smaller than the block length, rendering such optimizations unnecessary. 

\subsection{Sparse-dLLM}

We propose Sparse-dLLM, the first training-free framework to integrate a dynamic bidirectional cache eviction with sparse attention for dLLMs. Specifically, our method introduces two main strategies to manage the KV cache: (1) Dynamic bidirectional cache eviction, which leverages the temporal consistency in token saliency by using attention-aware sparse patterns to dynamically evict low-importance KV entries from both the prefix and suffix tokens (i.e., tokens preceding and succeeding the current block). (2) Delayed cache updates, where cache updates are intentionally delayed by one step to improve stability. The cache is fully cleared and refreshed when moving to a new decoding block.

\subsubsection{Dynamic Bidirectional Cache Eviction}
In contrast to auto-regressive LLMs that only sparsify prefix tokens, Sparse-dLLM widens the scope of cache eviction, targeting tokens from both the prefix (those preceding the current block) and the suffix (those succeeding the current block). Let $b$ denote the block length, $p$ denote the length of the prompt, and $o \in \left[ p, L \right)$ denote the positional offset of the first token in the current block. Then the candidate set $\mathbf{K}_{f}$, $\mathbf{V}_{f}$ from the KV states outside the current block are:
\begin{equation*}
\mathbf{K}_{f} = \text{Concat}[\mathbf{K}_{:o}, \mathbf{K}_{o+b:}],\ \ \mathbf{V}_{f} = \text{Concat}[\mathbf{V}_{:o}, \mathbf{V}_{o+b:}] 
\end{equation*}

Through observations from Figure~\ref{fig_sparsity}, we identify consistency in token saliency across queries and steps. This insight enables us to directly sparsify the cache by computing attention scores between the current block's query states and the candidate K states. For the current block's query states $\mathbf{Q}_{b}$, the attention scores are computed as: 

\begin{equation*}
\mathbf{A} = \frac{\mathbf{Q}_{b}\mathbf{K}_{f}^T}{\sqrt{d_k}}.
\end{equation*}

Drawing inspiration from SnapKV~\citep{li2025snapkv}, we incorporate max pooling operation to aggregate local information. This design prevents potential performance degradation caused by incomplete data when only partial details are preserved after dynamic cache eviction. Let $r \in [0,1]$ represent the retention ratio, $s$ represent the kernel size for max pooling. We can derive the indices of pivotal tokens through top-$k$ selection, where $k=(L-b)\times r$:
\begin{equation*}
\text{Indices} = \text{top-}k(\text{MaxPool}(\mathbf{A})).
\end{equation*}

The final KV cache $\mathbf{K}_{c}$, $\mathbf{V}_{c}$ is then constructed as:
\begin{equation*}
\mathbf{K}_{c} = \mathbf{K}_{f}[\text{Indices}],\ \ \mathbf{V}_{c} = \mathbf{V}_{f}[\text{Indices}]
\end{equation*}

Through its dynamic bidirectional cache eviction strategy, Sparse-dLLM effectively reduces the number of KV cache entries, which in turn decreases memory consumption while boosting inference throughput.

\subsubsection{Delayed Cache Updates}

\begin{figure}[!tb]
\centering
\includegraphics[width=0.85\linewidth]{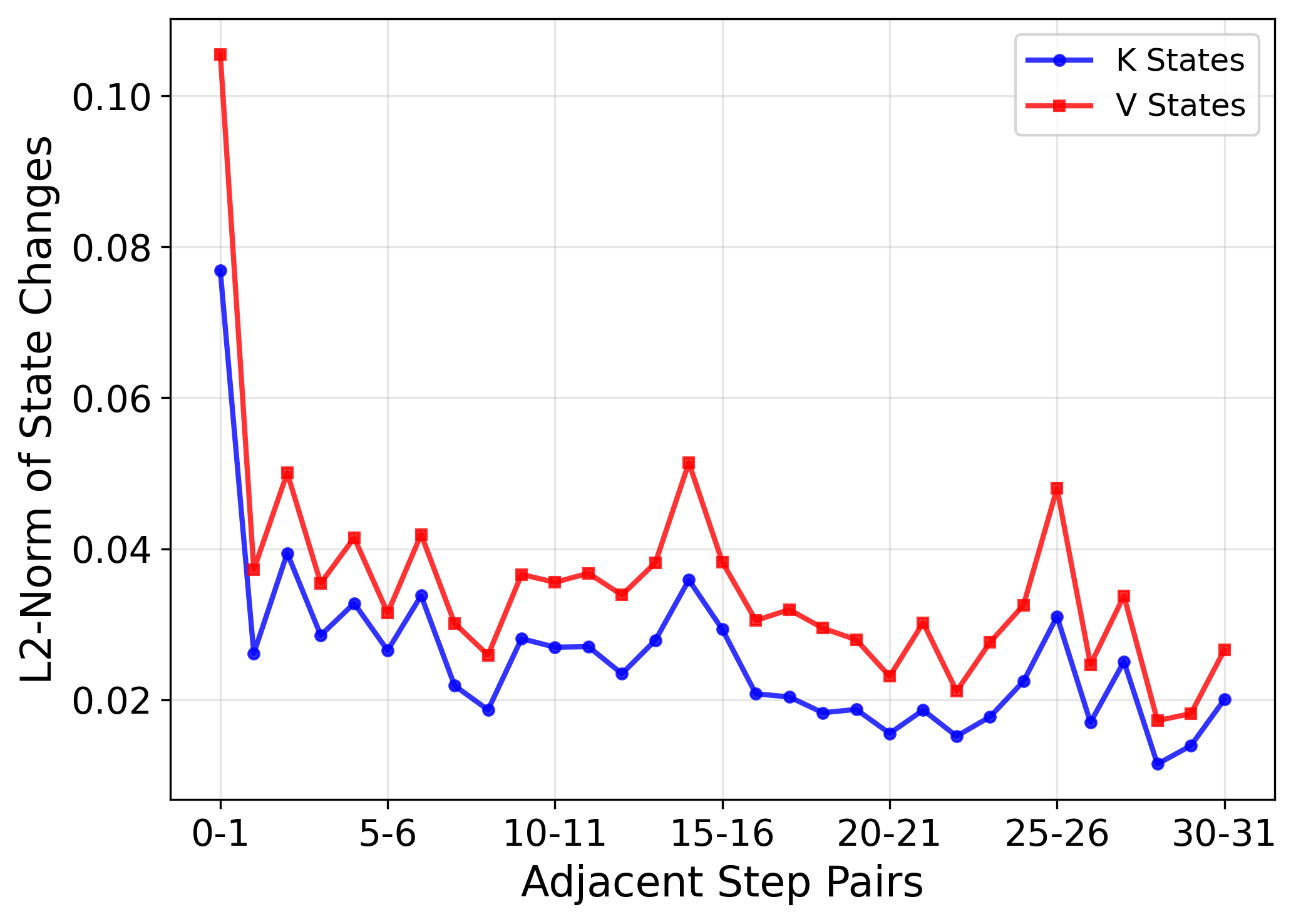}
\caption{L2-norm of KV state changes outside the current decoding block between adjacent step pairs.}
\label{kv_states_loss_for_delay_step}
\end{figure}

Furthermore, by observing the L2-norm of KV state changes outside the current decoding block between adjacent steps, as illustrated in Figure~\ref{kv_states_loss_for_delay_step}, it can be noted that the variation in KV states is relatively significant between step 0 and step 1. This observation suggests that the KV states intended for caching may not yet have stabilized at step 0 of the decoding block. Therefore, we delay the KV cache updates by one step upon decoding each block to mitigate early-stage instability in cached KV states.

\begin{table*}[!ht]
\tabcolsep=0.1cm
\centering
\small
\begin{tabular}{lccccccccccccccc}
\toprule
~ & \makecell{MMLU} & & \makecell{ARC-C} & & \makecell{PIQA} & & \makecell{GPQA} & & \makecell{GSM8k} & & \makecell{Math} & & \makecell{HE} & & Avg. \\ 
\midrule
\midrule
\textbf{LLaDA-8B-Instruct}& 60.60 & & \textbf{88.47} & & 83.62 & & 32.83 & & 78.39 & & \underline{36.02} & & 34.76 & & 59.24 \\
\ \ \ \ Throughput (TPS, $\uparrow$) & 9.48 & 1.0$\times$ & 21.08 & 1.0$\times$ & 21.74 & 1.0$\times$ & 6.30 & 1.0$\times$ & 4.57 & 1.0$\times$ & 8.52 & 1.0$\times$ & 12.53 & 1.0$\times$ & 1.0$\times$ \\
\ \ \ \ Memory (GB, $\downarrow$) & \textbf{15.54} & \textbf{1.0$\times$} & \textbf{15.18} & \textbf{1.0$\times$} & \textbf{15.17} & \textbf{1.0$\times$} & \textbf{15.86} & \textbf{1.0$\times$} & \textbf{16.18} & \textbf{1.0$\times$} & \textbf{15.63} & \textbf{1.0$\times$} & \textbf{15.39} & \textbf{1.0$\times$} & \textbf{1.0$\times$} \\
\midrule
+ dLLM-Cache & \underline{61.40} & & 87.46 & & 83.62 & & 32.83 & & \underline{78.92} & & \textbf{36.56} & & \textbf{37.80} & & 59.80 \\ 
\ \ \ \ Throughput (TPS, $\uparrow$) & \underline{21.43} & \underline{2.3$\times$} & \underline{24.03} & \underline{1.1$\times$} & \underline{23.64} & \underline{1.1$\times$} & 19.68 & 3.1$\times$ & 18.03 & 3.9$\times$ & \underline{21.11} & \underline{2.5$\times$} & \underline{21.64} & \underline{1.7$\times$} & \underline{2.3$\times$} \\
\ \ \ \ Memory (GB, $\downarrow$) & 16.61 & 1.1$\times$ & 15.82 & 1.0$\times$ & 15.80 & 1.0$\times$ & 17.19 & 1.1$\times$ & 17.85 & 1.1$\times$ & 16.74 & 1.1$\times$ & 16.65 & 1.1$\times$ & 1.1$\times$ \\
\midrule
+ dKV-Cache & 60.87 & & 87.80 & & 83.73 & & \textbf{35.35} & & \textbf{79.30} & & 35.46 & & 37.20 & & \textbf{59.96} \\ 
\ \ \ \ Throughput (TPS, $\uparrow$) & 14.34 & 1.5$\times$ & 18.28 & 0.9$\times$ & 18.34 & 0.8$\times$ & 10.95 & 1.7$\times$ & 8.89 & 1.9$\times$ & 13.46 & 1.6$\times$ & 14.40 & 1.1$\times$ & 1.4$\times$ \\
\ \ \ \ Memory (GB, $\downarrow$) & 17.88 & 1.2$\times$ & 16.10 & 1.1$\times$ & 16.05 & 1.1$\times$ & 19.46 & 1.2$\times$ & 21.08 & 1.3$\times$ & 18.34 & 1.2$\times$ & 17.17 & 1.1$\times$ & 1.2$\times$ \\
\midrule
+ Fast-dLLM & \textbf{61.43} & & 87.80 & & \underline{83.79} & & 32.83 & & 75.89 & & 33.78 & & 36.59 & & 58.87 \\ 
\ \ \ \ Throughput (TPS, $\uparrow$) & 20.51 & 2.2$\times$ & 21.63 & 1.0$\times$ & 21.99 & 1.0$\times$ & \underline{19.82} & \underline{3.1$\times$} & \underline{19.11} & \underline{4.2$\times$} & 20.72 & 2.4$\times$ & 21.50 & 1.7$\times$ & 2.2$\times$ \\
\ \ \ \ Memory (GB, $\downarrow$) & 17.13 & 1.1$\times$ & 15.81 & 1.0$\times$ & 15.77 & 1.0$\times$ & 18.29 & 1.2$\times$ & 19.48 & 1.2$\times$ & 17.47 & 1.1$\times$ & 16.60 & 1.1$\times$ & 1.1$\times$ \\
\midrule
+ Sparse-dLLM (ours) & 61.01 & & \textbf{88.47} & & \textbf{84.44} & & \textbf{35.35} & & 77.56 & & 34.42 & & \textbf{37.80} & & \underline{59.86} \\
\ \ \ \ Throughput (TPS, $\uparrow$) & \textbf{31.89} & \textbf{3.4$\times$} & \textbf{36.85} & \textbf{1.7$\times$} & \textbf{37.05} & \textbf{1.7$\times$} & \textbf{28.82} & \textbf{4.6$\times$} & \textbf{26.45} & \textbf{5.8$\times$} & \textbf{31.72} & \textbf{3.7$\times$} & \textbf{34.39} & \textbf{2.7$\times$} & \textbf{3.4$\times$} \\
\ \ \ \ Memory (GB, $\downarrow$) & \underline{15.73} & \underline{1.0$\times$} & \underline{15.26} & \underline{1.0$\times$} & \underline{15.24} & \underline{1.0$\times$} & \underline{16.16} & \underline{1.0$\times$} & \underline{16.60} & \underline{1.0$\times$} & \underline{15.86} & \underline{1.0$\times$} & \underline{15.54} & \underline{1.0$\times$} & \underline{1.0$\times$} \\
\midrule
\midrule
\textbf{LLaDA-1.5} & 61.05 & & \textbf{88.47} & & 83.79 & & \underline{33.33} & & 81.35 & & \underline{38.04} & & \textbf{40.24} & & 60.90 \\
\ \ \ \ Throughput (TPS, $\uparrow$) & 9.46 & 1.0$\times$ & 21.10 & 1.0$\times$ & 21.75 & 1.0$\times$ & 6.30 & 1.0$\times$ & 4.57 & 1.0$\times$ & 8.52 & 1.0$\times$ & 12.52 & 1.0$\times$ & 1.0$\times$ \\
\ \ \ \ Memory (GB, $\downarrow$) & \textbf{15.54} & \textbf{1.0$\times$} & \textbf{15.18} & \textbf{1.0$\times$} & \textbf{15.17} & \textbf{1.0$\times$} & \textbf{15.86} & \textbf{1.0$\times$} & \textbf{16.18} & \textbf{1.0$\times$} & \textbf{15.63} & \textbf{1.0$\times$} & \textbf{15.39} & \textbf{1.0$\times$} & \textbf{1.0$\times$} \\
\midrule
+ dLLM-Cache & \underline{61.41} & & \textbf{88.47} & & 83.62 & & 32.83 & & \underline{81.65} & & 37.04 & & 37.80 & & 60.40 \\
\ \ \ \ Throughput (TPS, $\uparrow$)  & \underline{21.40} & \underline{2.3$\times$} & \underline{23.95} & \underline{1.1$\times$} & \underline{24.85} & \underline{1.1$\times$} & \underline{20.41} & \underline{3.2$\times$} & 18.57 & 4.1$\times$ & \underline{21.72} & \underline{2.6$\times$} & 21.23 & 1.7$\times$ & \underline{2.3$\times$} \\
\ \ \ \ Memory (GB, $\downarrow$) & 16.61 & 1.1$\times$ & 15.83 & 1.0$\times$ & 15.80 & 1.0$\times$ & 17.19 & 1.1$\times$ & 17.85 & 1.1$\times$ & 16.74 & 1.1$\times$ & 16.65 & 1.1$\times$ & 1.1$\times$ \\
\midrule
+ dKV-Cache & 61.34 & & 88.14 & & \underline{84.28} & & \underline{33.33} & & \textbf{82.34} & & \textbf{38.08} & & \underline{39.63} & & \textbf{61.02} \\
\ \ \ \ Throughput (TPS, $\uparrow$)  & 14.29 & 1.5$\times$ & 18.26 & 0.9$\times$ & 18.41 & 0.8$\times$ & 10.98 & 1.7$\times$ & 8.91 & 2.0$\times$ & 13.62 & 1.6$\times$ & 14.40 & 1.2$\times$ & 1.4$\times$ \\
\ \ \ \ Memory (GB, $\downarrow$) & 17.88 & 1.2$\times$ & 16.10 & 1.1$\times$ & 16.05 & 1.1$\times$ & 19.46 & 1.2$\times$ & 21.08 & 1.3$\times$ & 18.34 & 1.2$\times$ & 17.17 & 1.1$\times$ & 1.2$\times$ \\
\midrule
+ Fast-dLLM & \textbf{61.57} & & 88.14 & & 83.95 & & 31.82 & & 80.82 & & 36.60 & & 36.59 & & 59.93 \\
\ \ \ \ Throughput (TPS, $\uparrow$)  & 20.74 & 2.2$\times$ & 21.89 & 1.0$\times$ & 21.67 & 1.0$\times$ & 20.09 & 3.2$\times$ & \underline{19.60} & \underline{4.3$\times$} & 20.59 & 2.4$\times$ & \underline{21.36} & \underline{1.7$\times$} & 2.3$\times$ \\
\ \ \ \ Memory (GB, $\downarrow$) & 17.13 & 1.1$\times$ & 15.81 & 1.0$\times$ & 15.77 & 1.0$\times$ & 18.29 & 1.2$\times$ & 19.48 & 1.2$\times$ & 17.47 & 1.1$\times$ & 16.60 & 1.1$\times$ & 1.1$\times$ \\
\midrule
+ Sparse-dLLM (ours) & 61.37 & & 88.14 & & \textbf{84.71} & & \textbf{34.34} & & 81.43 & & 37.32 & & \underline{39.63} & & \underline{60.99} \\ 
\ \ \ \ Throughput (TPS, $\uparrow$) & \textbf{32.05} & \textbf{3.4$\times$} & \textbf{36.78} & \textbf{1.7$\times$} & \textbf{36.96} & \textbf{1.7$\times$} & \textbf{29.06} & \textbf{4.6$\times$} & \textbf{26.21} & \textbf{5.7$\times$} & \textbf{31.87} & \textbf{3.7$\times$} & \textbf{34.18} & \textbf{2.7$\times$} & \textbf{3.4$\times$} \\
\ \ \ \ Memory (GB, $\downarrow$) & \underline{15.73} & \underline{1.0$\times$} & \underline{15.26} & \underline{1.0$\times$} & \underline{15.24} & \underline{1.0$\times$} & \underline{16.16} & \underline{1.0$\times$} & \underline{16.60} & \underline{1.0$\times$} & \underline{15.86} & \underline{1.0$\times$} & \underline{15.54} & \underline{1.0$\times$} & \underline{1.0$\times$} \\
\bottomrule
\end{tabular}
\caption{Comprehensive benchmark results on LLaDA-8B-Instruct~\citep{nie2025large} and LLaDA-1.5~\citep{zhu2025llada}. Each cell presents the accuracy, decoding throughput in tokens per second and peak memory cost in GB with relative efficiency to the pre-trained model. Best values in bold, suboptimal values underlined.}
\label{tab-llada_exp}
\end{table*}

\begin{table*}[!ht]
\tabcolsep=0.1cm
\centering
\small
\begin{tabular}{lccccccccccccccc}
\toprule
~ & \makecell{MMLU} & & \makecell{ARC-c} & & \makecell{PIQA} & & \makecell{GPQA} & & \makecell{GSM8k} & & \makecell{Math} & & \makecell{HE} & & Avg. \\ 
\midrule
\midrule
\textbf{Dream-v0-7B-Base} & \textbf{72.96} & & 82.71 & & 81.18 & & \textbf{32.83} & & 70.74 & & \underline{20.78} & & \textbf{53.05} & & \underline{59.18} \\ 
\ \ \ \ Throughput (TPS, $\uparrow$)  & 9.88 & 1.0$\times$ & 19.60 & 1.0$\times$ & 20.15 & 1.0$\times$ & 6.33 & 1.0$\times$ & 7.38 & 1.0$\times$ & 8.93 & 1.0$\times$ & 11.80 & 1.0$\times$ & 1.0$\times$ \\
\ \ \ \ Memory (GB, $\downarrow$) & \underline{15.64} & \underline{1.0$\times$} & \underline{15.49} & \underline{1.0$\times$} & \underline{15.49} & \underline{1.0$\times$} & \underline{15.77} & \underline{1.0$\times$} & \underline{15.73} & \underline{1.0$\times$} & \underline{15.67} & \underline{1.0$\times$} & \underline{16.73} & \underline{1.0$\times$} & \underline{1.0$\times$} \\
\midrule
+ dLLM-Cache & \underline{72.87} & & 85.08 & & 81.12 & & 31.31 & & 72.55 & & 20.40 & & \underline{50.61} & & 59.13 \\ 
\ \ \ \ Throughput (TPS, $\uparrow$)  & 12.68 &  1.3$\times$ & 19.33 &  1.0$\times$ & 19.75 &  1.0$\times$ & 8.72 &  1.4$\times$ & 9.87 &  1.3$\times$ & 12.16 &  1.4$\times$ & 14.24 &  1.2$\times$ & 1.2$\times$ \\ 
\ \ \ \ Memory (GB, $\downarrow$) & 16.37 &  1.0$\times$ & 15.79 &  1.0$\times$ & 15.77 &  1.0$\times$ & 16.93 &  1.1$\times$ & 16.76 &  1.1$\times$ & 16.50 &  1.1$\times$ & 17.28 &  1.0$\times$ & 1.0$\times$ \\ 
\midrule
+ dKV-Cache & 72.77 & & 82.03 & & \underline{81.39} & & \textbf{32.83} & & 69.90 & & 20.06 & & 45.73 & & 57.82 \\ 
\ \ \ \ Throughput (TPS, $\uparrow$)  & 13.90 &  1.4$\times$ & 19.65 &  1.0$\times$ & 19.86 &  1.0$\times$ & 11.03 &  1.7$\times$ & 11.91 &  1.6$\times$ & 13.43 &  1.5$\times$ & 13.14 &  1.1$\times$ & 1.3$\times$ \\ 
\ \ \ \ Memory (GB, $\downarrow$) & 15.92 &  1.0$\times$ & 15.60 &  1.0$\times$ & 15.59 &  1.0$\times$ & 16.23 &  1.0$\times$ & 16.14 &  1.0$\times$ & 16.00 &  1.0$\times$ & 16.95 &  1.0$\times$ & 1.0$\times$ \\ 
\midrule
+ Fast-dLLM & 72.69 & & \textbf{86.78} & & \textbf{82.86} & & 31.31 & & \underline{73.09} & & 19.90 & & 41.46 & & 58.30 \\ 
\ \ \ \ Throughput (TPS, $\uparrow$)  & \underline{25.72} &  \underline{2.6$\times$} & \underline{27.18} &  \underline{1.4$\times$} & \underline{26.78} &  \underline{1.3$\times$} & \underline{24.26} &  \underline{3.8$\times$} & \underline{24.78} &  \underline{3.4$\times$} & \underline{25.39} &  \underline{2.8$\times$} & \underline{26.19} &  \underline{2.2$\times$} & \underline{2.5$\times$} \\ 
\ \ \ \ Memory (GB, $\downarrow$) & 18.32 &  1.2$\times$ & 15.85 &  1.0$\times$ & 15.77 &  1.0$\times$ & 20.69 &  1.3$\times$ & 19.95 &  1.3$\times$ & 18.94 &  1.2$\times$ & 17.31 &  1.0$\times$ & 1.1$\times$ \\ 
\midrule
+ Sparse-dLLM (ours) & 72.61 & & \textbf{86.78} & & \underline{81.39} & & 30.81 & & \textbf{74.15} & & \textbf{23.60} & & 45.12 & & \textbf{59.21} \\ 
\ \ \ \ Throughput (TPS, $\uparrow$)  & \textbf{36.97} &  \textbf{3.7$\times$} & \textbf{42.00} &  \textbf{2.1$\times$} & \textbf{42.27} &  \textbf{2.1$\times$} & \textbf{32.71} &  \textbf{5.2$\times$} & \textbf{34.28} &  \textbf{4.6$\times$} & \textbf{36.56} &  \textbf{4.1$\times$} & \textbf{38.91} &  \textbf{3.3$\times$} & \textbf{3.6$\times$} \\ 
\ \ \ \ Memory (GB, $\downarrow$) & \textbf{14.74} &  \textbf{0.9$\times$} & \textbf{14.52} &  \textbf{0.9$\times$} & \textbf{14.52} &  \textbf{0.9$\times$} & \textbf{15.03} &  \textbf{1.0$\times$} & \textbf{14.94} &  \textbf{0.9$\times$} & \textbf{14.81} &  \textbf{0.9$\times$} & \textbf{14.62} &  \textbf{0.9$\times$} & \textbf{0.9$\times$} \\ 
\midrule
\midrule
\textbf{Dream-v0-7B-Instruct} & \underline{72.42} & & \textbf{90.17} & & 88.25 & & \underline{34.85} & & 76.57 & & 39.38 & & \underline{57.93} & & \textbf{65.65} \\ 
\ \ \ \ Throughput (TPS, $\uparrow$)  & 9.61 & 1.0$\times$ & 19.23 & 1.0$\times$ & 19.65 & 1.0$\times$ & 6.29 & 1.0$\times$ & 7.27 & 1.0$\times$ & 8.93 & 1.0$\times$ & 11.41 & 1.0$\times$ & 1.0$\times$ \\
\ \ \ \ Memory (GB, $\downarrow$) & \underline{15.64} & \underline{1.0$\times$} & \underline{15.50} & \underline{1.0$\times$} & \underline{15.49} & \underline{1.0$\times$} & \underline{15.78} & \underline{1.0$\times$} & \underline{15.73} & \underline{1.0$\times$} & \underline{15.68} & \underline{1.0$\times$} & \underline{16.74} & \underline{1.0$\times$} & \underline{1.0$\times$} \\
\midrule
+ dLLM-Cache & \textbf{72.55} & & \textbf{90.17} & & \textbf{88.79} & & \underline{34.85} & & 75.74 & & 37.44 & & \textbf{59.15} & & \underline{65.53} \\ 
\ \ \ \ Throughput (TPS, $\uparrow$)  & 12.48 &  1.3$\times$ & 19.54 &  1.0$\times$ & 19.66 &  1.0$\times$ & 8.66 &  1.4$\times$ & 9.79 &  1.3$\times$ & 12.11 &  1.4$\times$ & 13.97 &  1.2$\times$ & 1.2$\times$ \\ 
\ \ \ \ Memory (GB, $\downarrow$) & 16.41 &  1.0$\times$ & 15.80 &  1.0$\times$ & 15.79 &  1.0$\times$ & 16.96 &  1.1$\times$ & 16.78 &  1.1$\times$ & 16.52 &  1.1$\times$ & 17.30 &  1.0$\times$ & 1.0$\times$ \\ 
\midrule
+ dKV-Cache & 72.37 & & \textbf{90.17} & & \underline{88.36} & & 33.33 & & 77.48 & & 37.94 & & 56.10 & & 65.11 \\ 
\ \ \ \ Throughput (TPS, $\uparrow$)  & 14.14 &  1.5$\times$ & 19.44 &  1.0$\times$ & 19.65 &  1.0$\times$ & 10.82 &  1.7$\times$ & 11.97 &  1.6$\times$ & 13.33 &  1.5$\times$ & 12.97 &  1.1$\times$ & 1.4$\times$ \\ 
\ \ \ \ Memory (GB, $\downarrow$) & 15.93 &  1.0$\times$ & 15.61 &  1.0$\times$ & 15.60 &  1.0$\times$ & 16.24 &  1.0$\times$ & 16.15 &  1.0$\times$ & 16.01 &  1.0$\times$ & 16.96 &  1.0$\times$ & 1.0$\times$ \\ 
\midrule
+ Fast-dLLM & 70.81 & & 89.83 & & 86.02 & & 31.82 & & \underline{77.79} & & \underline{39.82} & & 52.44 & & 64.08 \\ 
\ \ \ \ Throughput (TPS, $\uparrow$)  & \underline{25.37} &  \underline{2.6$\times$} & \underline{27.66} &  \underline{1.4$\times$} & \underline{27.75} &  \underline{1.4$\times$} & \underline{24.54} & \underline{3.9$\times$} & \underline{24.57} &  \underline{3.4$\times$} & \underline{25.43} &  \underline{2.8$\times$} & \underline{26.15} &  \underline{2.3$\times$} & \underline{2.6$\times$} \\ 
\ \ \ \ Memory (GB, $\downarrow$) & 18.41 &  1.2$\times$ & 15.94 &  1.0$\times$ & 15.86 &  1.0$\times$ & 
20.78 &  1.3$\times$ & 20.03 &  1.3$\times$ & 19.02 &  1.2$\times$ & 17.39 &  1.0$\times$ & 1.2$\times$ \\ 
\midrule
+ Sparse-dLLM (ours) & 70.94 & & 88.47 & & 86.62 & & \textbf{35.86} & & \textbf{78.17} & & \textbf{40.48} & & 50.00 & & 64.36 \\ 
\ \ \ \ Throughput (TPS, $\uparrow$)  & \textbf{36.89} &  \textbf{3.8$\times$} & \textbf{42.01} &  \textbf{2.2$\times$} & \textbf{42.11} &  \textbf{2.1$\times$} & \textbf{32.63} &  \textbf{5.2$\times$} & \textbf{34.10} &  \textbf{4.7$\times$} & \textbf{36.72} &  \textbf{4.1$\times$} & \textbf{38.59} &  \textbf{3.4$\times$} & \textbf{3.6$\times$} \\ 
\ \ \ \ Memory (GB, $\downarrow$) & \textbf{14.75} &  \textbf{0.9$\times$} & \textbf{14.53} &  \textbf{0.9$\times$} & \textbf{14.52} &  \textbf{0.9$\times$} & 
\textbf{15.04} &  \textbf{1.0$\times$} & \textbf{14.95} &  \textbf{1.0$\times$} & \textbf{14.83} &  \textbf{0.9$\times$} & \textbf{14.62} &  \textbf{0.9$\times$} & \textbf{0.9$\times$} \\ 
\bottomrule
\end{tabular}
\caption{Comprehensive benchmark results on Dream-v0-7B and Dream-v0-7B-Instruct~\citep{dream2025}. Each cell presents the accuracy, decoding throughput in tokens per second and peak memory cost in GB with relative efficiency. Best values in bold, suboptimal values underlined.}
\label{tab-dream_exp}
\end{table*}

\section{Experiment}

\subsection{Setup}

We conduct experiments on the existing dLLMs, including LLaDA-8B-Instruct~\citep{nie2025large}, LLaDA-1.5~\citep{zhu2025llada}, Dream-v0-7B-Base, and Dream-v0-7B-Instruct~\citep{dream2025}. By default, we set the block length to 32 and keep the unmasking strategy in the official code of LLaDA and Dream. For all models, we apply a fixed random seed 2025, retention ratio $r=0.5$, and kernel size $s=3$.  % series And of

The evaluation metrics comprise accuracy for benchmarks, throughput (measured in Tokens Per Second, TPS), and peak memory consumption (GB), providing a comprehensive assessment of the model's performance and efficiency. We use OpenCompass~\citep{2023opencompass} for validation. The benchmarks we use cover general tasks, science, mathematics, and code, including MMLU (5-shot)~\citep{hendrycks2020mmlu}, ARC-challenge (ARC-c, 0-shot)~\citep{clark2018arc}, PIQA (0-shot)~\citep{bisk2020piqa}, GPQA (5-shot)~\citep{rein2024gpqa}, GSM8k (4-shot)~\citep{cobbe2021gsm8k}, Math (4-shot)~\citep{hendrycks2021math}, and HumanEval (HE, 0-shot)~\citep{chen2021humaneval}. All experiments were performed on NVIDIA 4090 (48 GB) GPUs.

The evaluation was conducted for each benchmark as follows: For performance, we report the mean score over three independent trials. For efficiency, we report the average results generated from the same ten randomly sampled data instances used across all methods.

\subsection{Main Results}

The main results comparing the baseline, other methods and our proposed Sparse-dLLM on LLaDA and Dream are presented in Table~\ref{tab-llada_exp} and Table~\ref{tab-dream_exp}, respectively. These results demonstrate that our Sparse-dLLM method achieves the most significant throughput improvement while maintaining or even slightly enhancing performance, with nearly identical peak memory to vanilla dLLMs. 

Sparse-dLLM achieves a remarkable leap in throughput. When applied to the LLaDA-8B-Instruct model, Sparse-dLLM increases the throughput from 4.57 TPS to 26.45 TPS on the GSM8K dataset, achieving a 5.8$\times$ speedup. On the GPQA dataset, when applied to the Dream-v0-7B-Instruct model, Sparse-dLLM improves the throughput from 6.29 TPS to 32.63 TPS, yielding a 5.2$\times$ acceleration. Moreover, Sparse-dLLM consistently achieves the highest throughput across all model and benchmark configurations.

Sparse-dLLM maintains nearly identical memory costs compared to vanilla dLLMs. On LLaDA models, Sparse-dLLM's peak memory remains nearly identical to the baseline (within 0.5 GB), while other methods significantly increase memory consumption. More significantly, through the introduction of block-wise decoding on Dream, Sparse-dLLM's sampling process only requires the logits from the current block. Consequently, even with the introduction of KV cache, Sparse-dLLM's memory consumption remains lower than the baseline.

Sparse-dLLM achieves significant efficiency improvements while maintaining comparable performance. On LLaDA series models, Sparse-dLLM enhances inference efficiency and even slightly boosts accuracy across benchmarks. On the Dream-v0-7B-Base model, Sparse-dLLM demonstrates superior average performance compared to both the baseline and all other methods. Our Sparse-dLLM demonstrates inferior performance compared to the baseline and existing cache-based methods on the HumanEval benchmark on Dream. We conjecture that a complete context is more necessary for code tasks. 

\begin{figure}[!tb]
\centering
\begin{subfigure}[b]{0.48\linewidth}
    \centering
    \includegraphics[width=0.92\linewidth]{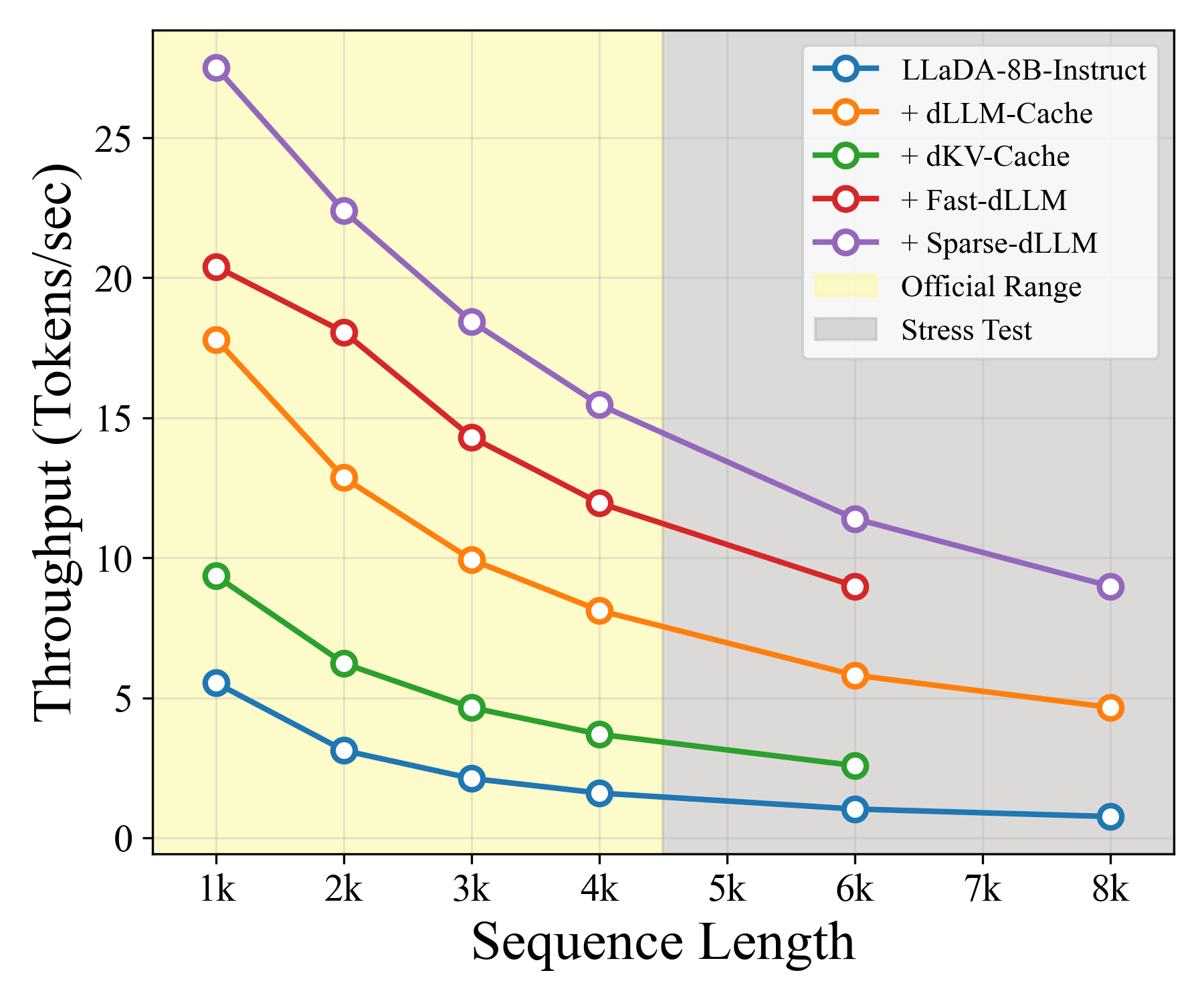}
    \caption{Throughput (LLaDA)}
    \label{diff_len_llada_tps}
\end{subfigure}
\hfill
\begin{subfigure}[b]{0.48\linewidth}
    \centering
    \includegraphics[width=0.92\linewidth]{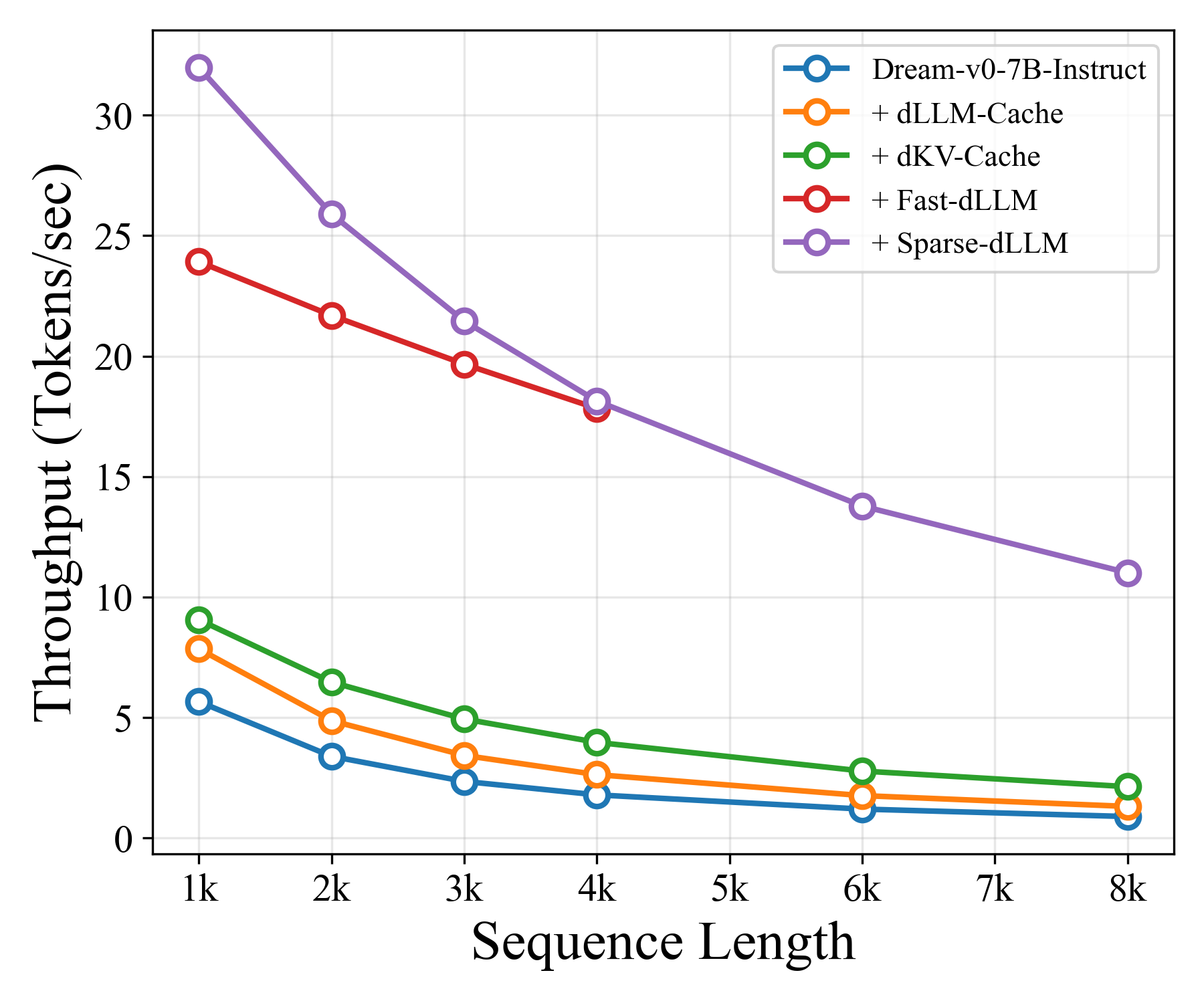}
    \caption{Throughput (Dream)}
    \label{diff_len_dream_tps}
\end{subfigure}
\\[0.9ex]
\begin{subfigure}[b]{0.48\linewidth}
    \centering
    \includegraphics[width=0.92\linewidth]{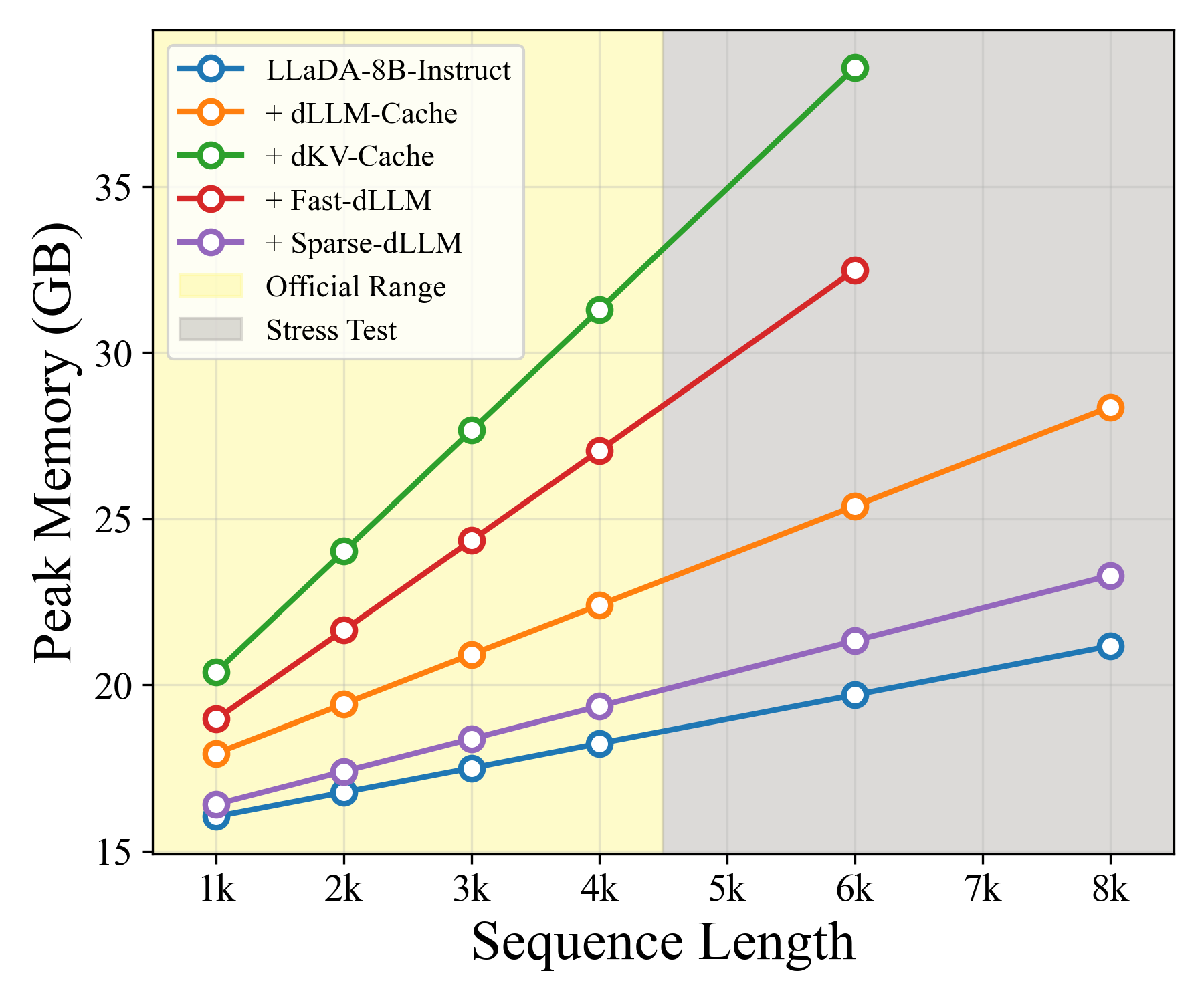}
    \caption{Peak Memory (LLaDA)}
    \label{diff_len_llada_mem}
\end{subfigure}
\hfill
\begin{subfigure}[b]{0.48\linewidth}
    \centering
    \includegraphics[width=0.92\linewidth]{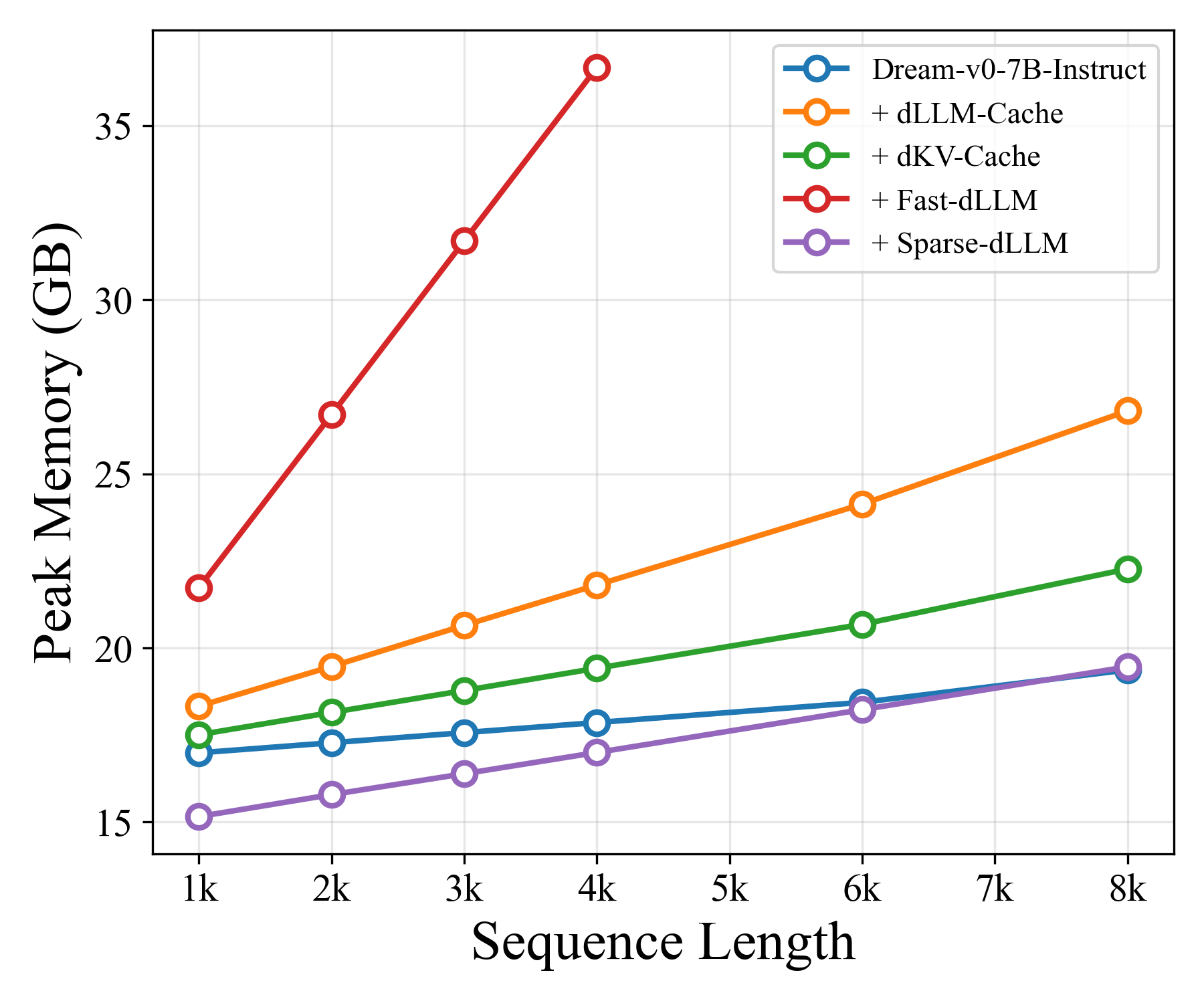}
    \caption{Peak Memory (Dream)}
    \label{diff_len_dream_mem}
\end{subfigure}
\caption{Results of efficiency comparison in different context lengths. The missing data points indicate that the combination of the model and method resulted in an OOM error on the NVIDIA 4090 (48 GB) GPU. We adopt LLaDA-8B-Instruct (LLaDA) and Dream-v0-7B-Instruct (Dream). \label{fig_diff_len}}
\end{figure}

\subsection{Long-Context Efficiency}

In Figure~\ref{fig_diff_len}, evaluation of different methods in processing short and long contexts (stress test) shows that Sparse-dLLM (purple line) exhibits the most comprehensive advantages. 

In terms of throughput, Sparse-dLLM consistently outperforms all other methods on both LLaDA and Dream models, achieving a remarkable speedup of up to 10$\times$ higher throughput than vanilla dLLMs at 4k sequence length. In contrast, although Fast-dLLM (red line) demonstrates competitive throughput at 4k sequence length on Dream, its memory consumption increases drastically with sequence length, eventually leading to Out-of-Memory (OOM) errors on an NVIDIA 4090 (48 GB) GPU when processing long context. Other methods, such as dLLM-Cache (orange line) and dKV-Cache (green line), show limited throughput improvements compared to the baseline model.

Regarding peak memory consumption, Sparse-dLLM exhibits exemplary performance. Its memory growth curve is notably flat, indicating a growth rate merely marginally above that of the baseline. In comparison to all other cache-based methods, it maintains a significantly lower peak memory. Overall, Sparse-dLLM successfully achieves a balance between high throughput and low memory consumption, confirming its superiority as an efficient and scalable solution for long-context processing.

\section{Ablations and Analysis}

\paragraph{N-Step Delayed Cache Updates}

We systematically examined the effects of the delay step (0-5) for cache updates during decoding a new block on LLaDA-8B-Instruct, as shown in Table~\ref{tab_delay_step}. The results demonstrate that increasing the delay step leads to progressively lower throughput. Interestingly, accuracy exhibits a non-monotonic pattern, decreasing when transitioning from 1-step to 2-step delay and reaching its peak at 3-step delay. Through a comprehensive analysis of this performance-efficiency trade-off, we conclude that 1-step delay represents the optimal setting, offering near-optimal accuracy while maintaining a near-maximum throughput.

\begin{table}[!t]
\centering
\tabcolsep=0.12cm
\small
\begin{tabular}{lcccccc}
\toprule
Delay Step & 0 & 1 & 2 & 3 & 4 & 5 \\
\midrule 
Accuracy (\%) & 86.1 & \underline{88.47} & 87.46 & \textbf{89.49} & \underline{88.47} & 88.14 \\
Throughput (TPS) & \textbf{36.89} & \underline{36.85} & 36.26 & 35.38 & 34.89 & 33.78 \\
\bottomrule
\end{tabular}
\caption{Ablation study of delay step on ARC-C benchmark.}
\label{tab_delay_step}
\end{table}

\paragraph{Sparsity Strategy}

\begin{table}[!t]
\centering
\tabcolsep=0.18cm
\small
\begin{tabular}{lcccc}
\toprule
~ & GSM8k & MATH & ARC-C & Avg. \\
\midrule 
\textbf{LLaDA-8B-Instruct} & \underline{78.39} & \textbf{36.02} & \textbf{88.47} & \textbf{67.63} \\
+ Sparse-dLLM & 77.56 & \underline{34.42} & \textbf{88.47} & \underline{66.82} \\
+ prefix-sparse & \textbf{78.62} & 34.18 & 83.39 & 65.40 \\
\midrule 
\textbf{Dream-v0-7B-Instruct} & 76.57 & 39.38 & \textbf{90.17} & 68.71 \\
+ Sparse-dLLM & \underline{78.17} & \textbf{40.48} & 88.47 & \textbf{69.04} \\
+ prefix-sparse & \textbf{78.24} & \underline{39.74} & \underline{89.15} & \textbf{69.04} \\
\bottomrule
\end{tabular}
\caption{Ablation on different sparsity strategies.}
\label{ablation_strategy}
\end{table}

To evaluate the effectiveness of our bidirectional sparsification, we conducted the ablation study presented in Table~\ref{ablation_strategy}. The results reveal that our Sparse-dLLM, which sparsifies KV states both preceding and succeeding the current block, outperforms the unidirectional prefix-sparse approach. While both strategies enhance performance on the Dream-v0-7B-Instruct model, our approach's advantage is particularly evident on the challenging MATH dataset. Furthermore, although both methods cause a marginal performance drop on the LLaDA-8B-Instruct model, our approach mitigates this degradation much more effectively. These findings underscore that bidirectional sparsification is a more effective strategy.

\paragraph{Hyperparameters: Retention Ratio and Kernel Size}

\begin{figure}[!tb]
\centering
\begin{subfigure}[b]{0.48\linewidth}
    \centering
    \includegraphics[width=0.92\linewidth]{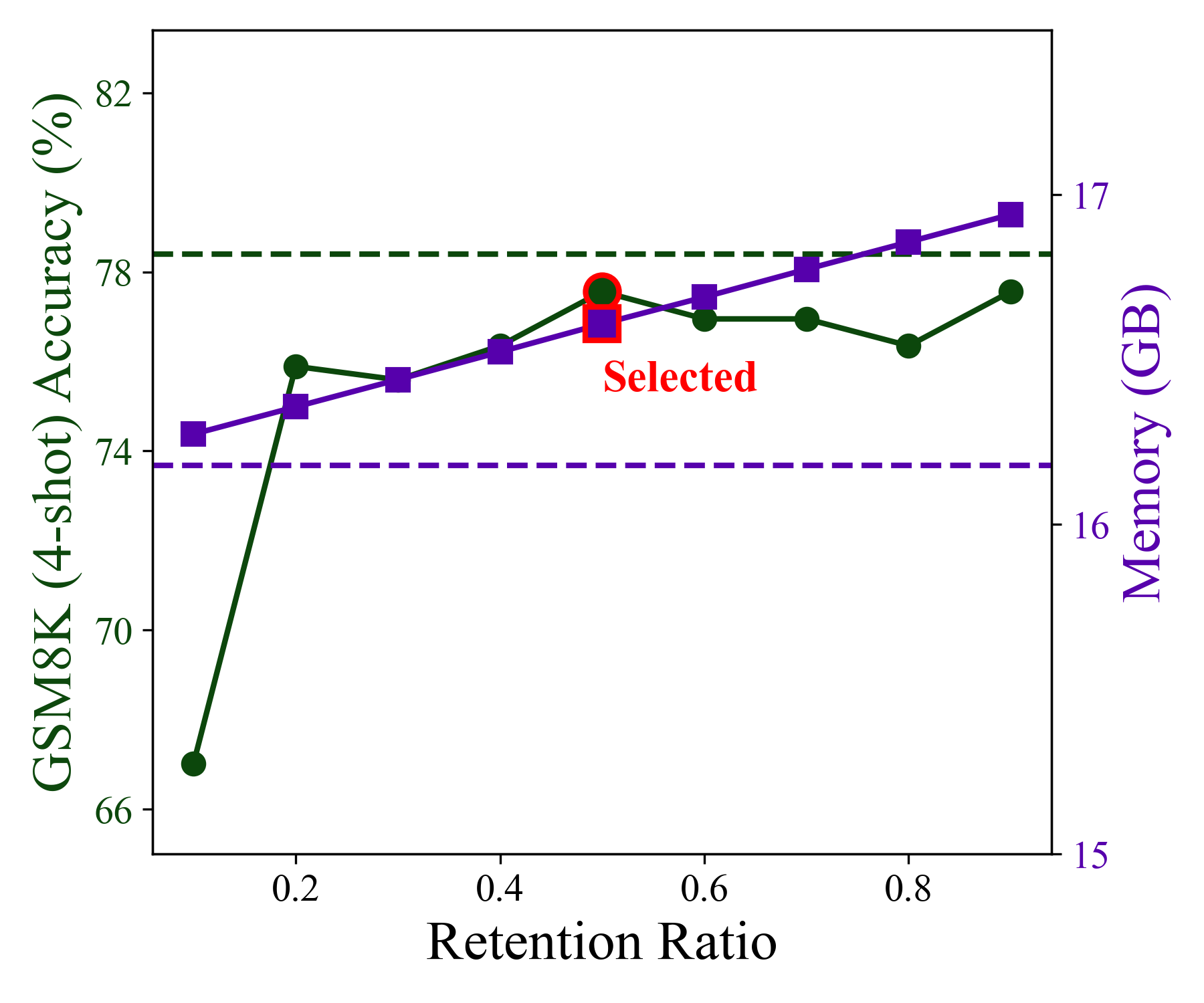}
    \caption{Retention Ratio (LLaDA)}
    \label{fig_llada_diff_kr}
\end{subfigure}
\hfill
\begin{subfigure}[b]{0.48\linewidth}
    \centering
    \includegraphics[width=0.92\linewidth]{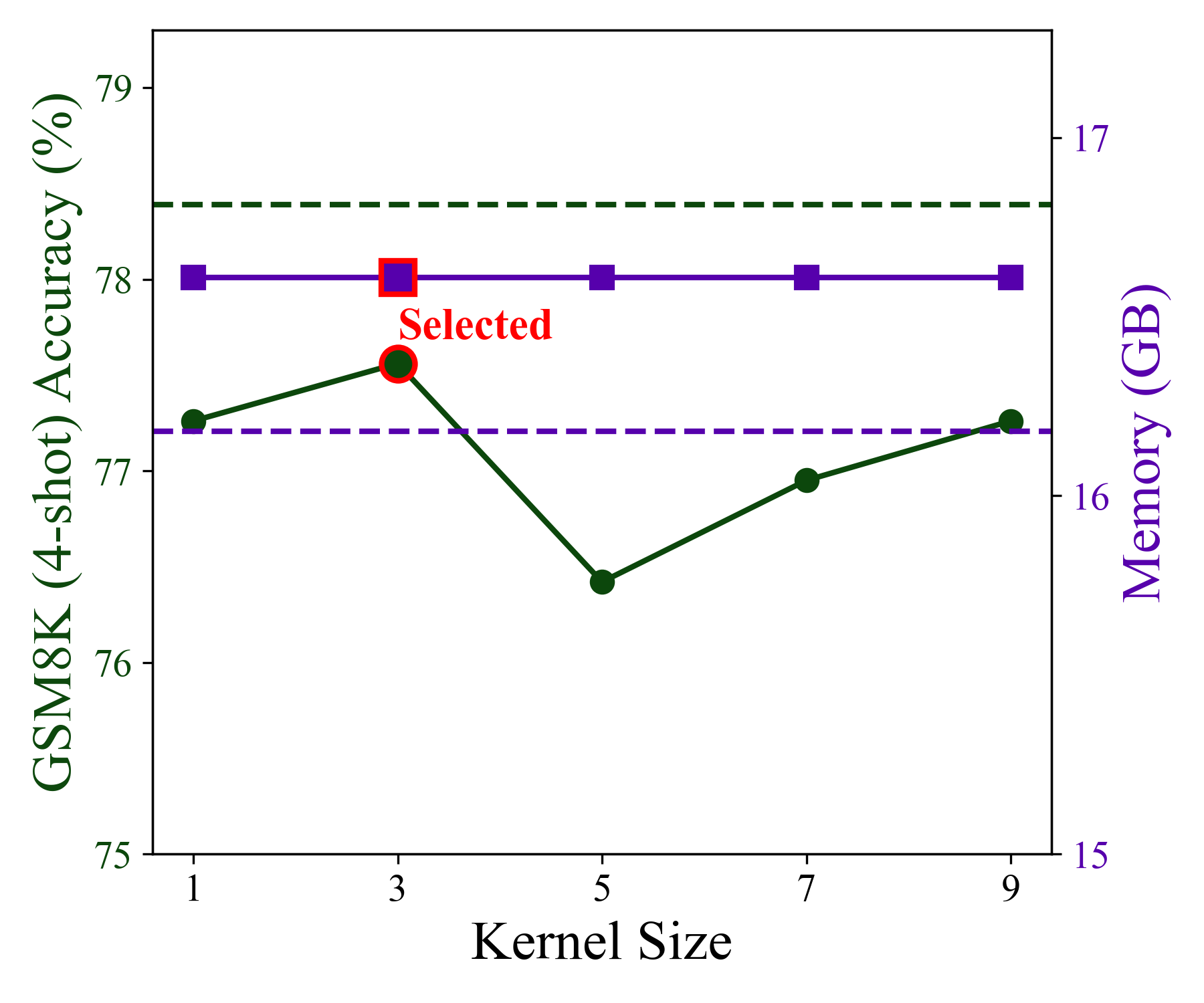}
    \caption{Kernel Size (LLaDA)}
    \label{fig_llada_diff_ks}
\end{subfigure}
\\[0.9ex]
\begin{subfigure}[b]{0.48\linewidth}
    \centering
    \includegraphics[width=0.92\linewidth]{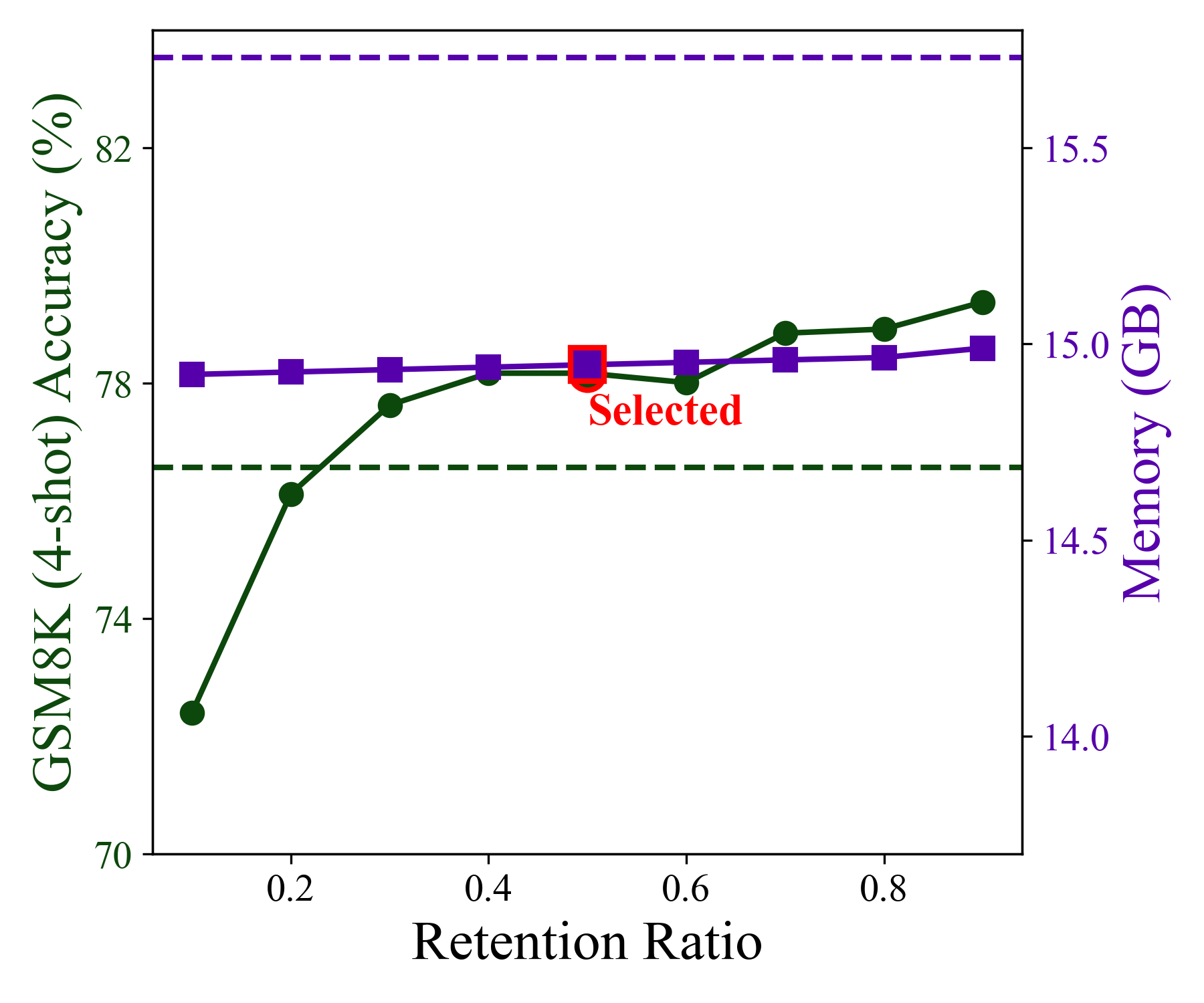}
    \caption{Retention Ratio (Dream)}
    \label{fig_dream_diff_kr}
\end{subfigure}
\hfill
\begin{subfigure}[b]{0.48\linewidth}
    \centering
    \includegraphics[width=0.92\linewidth]{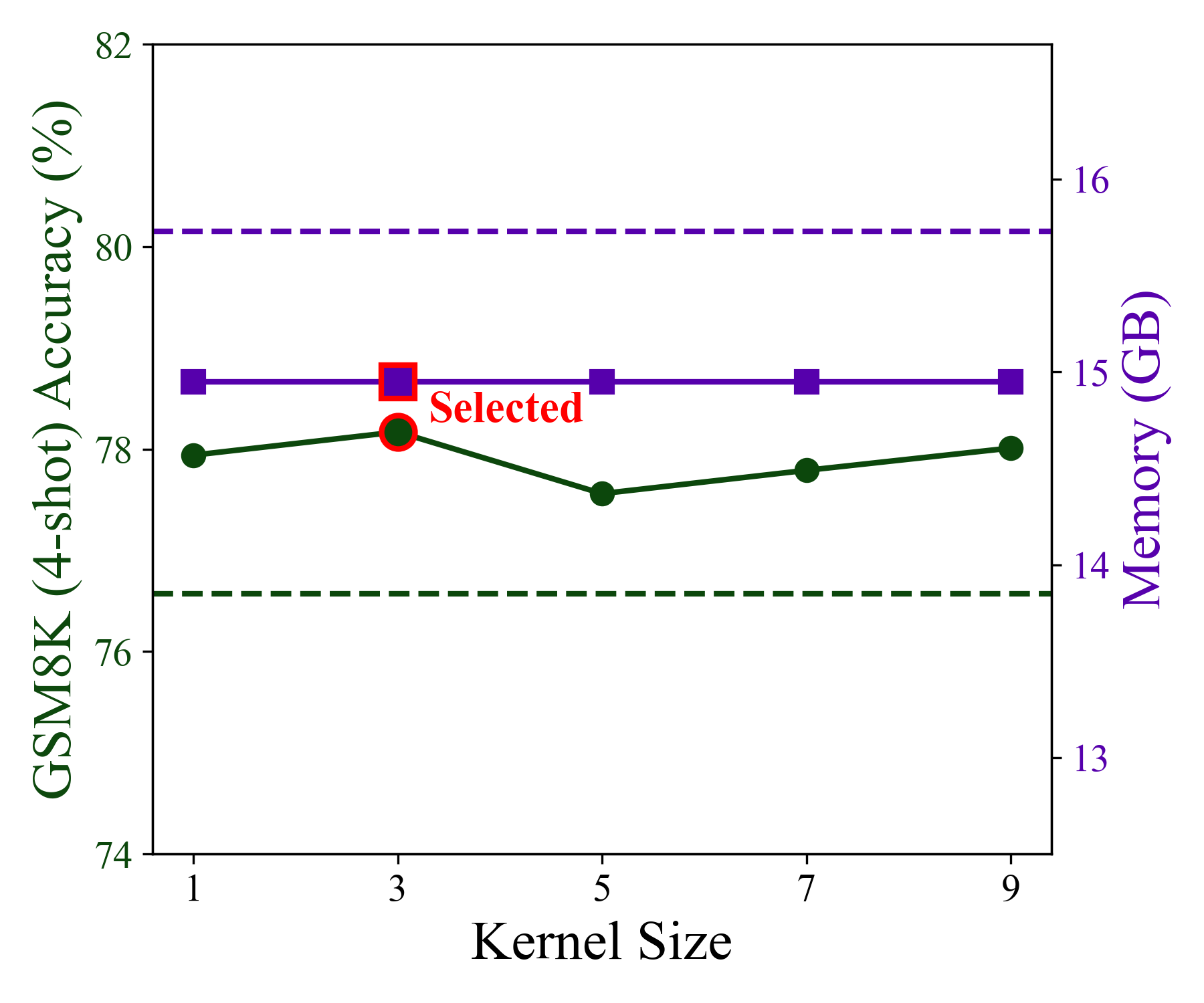}
    \caption{Kernel Size (Dream)}
    \label{fig_dream_diff_ks}
\end{subfigure}
\caption{Ablation on retention ratio and kernel size. The left vertical axis denotes the accuracy on GSM8K (4-shot), while the right vertical axis represents the peak memory. We adopt LLaDA-8B-Instruct (LLaDA) and Dream-v0-7B-Instruct (Dream). \label{fig_diff_kr_ks}}
\end{figure}

To determine the optimal hyperparameters for Sparse-dLLM, we conducted systematic ablation studies using the GSM8K (4-shot) benchmark accuracy and peak memory as our evaluation metrics, examining both retention ratio and kernel size configurations, as shown in Figure~\ref{fig_diff_kr_ks}. In Figures~\ref{fig_llada_diff_kr} and~\ref{fig_dream_diff_kr} with fixed kernel size $k=3$, increasing the retention ratio from 0.1 to 0.5 yields substantial accuracy improvements on both LLaDA and Dream. When increasing the retention ratio beyond 0.5, empirical results indicate that it leads to diminishing performance gains, with marginal improvements at some retention ratios and even slight degradation at others, while memory consumption continues to grow linearly. This saturation phenomenon suggests an optimal trade-off between model performance and computational efficiency at $r=0.5$. Figures~\ref{fig_llada_diff_ks} and~\ref{fig_dream_diff_ks} with fixed retention ratio $r=0.5$ reveal a clear performance peak at kernel size $k=3$ on both LLaDA and Dream, with smaller ($k=1$) or larger ($k\geq5$) kernels both degrading accuracy. Based on this empirical evidence, we establish $r=0.5$ and $k=3$ as the optimal hyperparameters for all main experiments.

\section{Conclusion}

We introduce Sparse-dLLM, the first training-free method to combine sparse attention with dynamic bidirectional cache eviction for dLLMs. Based on the key insight that attention in dLLMs is both sparse and consistent across decoding steps, our method dynamically evicts unimportant KV cache entries for both prefix and suffix tokens. Our approach achieves three key results: 1) Comparable performance on downstream tasks; 2) State-of-the-art acceleration, with up to 10$\times$ higher throughput than vanilla dLLMs; and 3) Optimal memory efficiency, with nearly identical memory costs to vanilla dLLMs.

\bibliography{aaai2026}


\appendix
\onecolumn

\section{Algorithmic Pseudocode}

\subsection{Cache Implementation for Sparse-dLLM}

Algorithm~\ref{alg:dynamic_cache_eviction} outlines the core cache implementation, dynamic bidirectional cache eviction, of our Sparse-dLLM. During cache updates, the Key-Value (KV) states excluding the current block are extracted. These states are then evaluated by computing their average attention scores using the query states from the current block. Subsequently, max pooling is applied to obtain importance scores. Based on the retention ratio $r$, pivotal tokens' KV states are selected and stored in the cache for subsequent reuse.

\begin{algorithm}[!ht]
\caption{Dynamic Bidirectional Cache Eviction}
\label{alg:dynamic_cache_eviction}
\begin{algorithmic}[1]
    \REQUIRE Layer id $l$, query states of the current block $\mathbf{Q}_{b}$, KV states $\mathbf{K}$ and $\mathbf{V}$ for the full sequence, offset of the current block $o$,  block length $b$, retention ratio $r$, kernel size $k$.
    \ENSURE Updated cache for layer $l$. 
    
    \STATE /* Extract KV states excluding the current block */
    \STATE $\mathbf{K}_{f} \leftarrow \text{Concat}(\mathbf{K}_{:o}, \mathbf{K}_{o+b:})$
    \STATE $\mathbf{V}_{f} \leftarrow \text{Concat}(\mathbf{V}_{:o}, \mathbf{V}_{o+b:})$

    \STATE /* Calculate attention scores */
    \STATE $\bar{q} \leftarrow \text{Mean}(\mathbf{Q}_{b})$ \hfill \COMMENT{Average query state in the block}
    \STATE $\mathbf{A} \leftarrow \text{MatMul}(\bar{q}, \mathbf{K}_{f}^T)$ \hfill \COMMENT{Calculate attention scores}

    \STATE /* Max pooling and select tokens */
    \STATE \parbox[t]{\linewidth}{
        $\mathbf{I} \leftarrow \text{MaxPool1D}(\mathbf{A}, \text{kernel\_size}=k, \text{padding}=k/2)$ \hfill \COMMENT{Calculate importance scores}
    }
    \STATE $n \leftarrow \lfloor |\mathbf{I}| \cdot r \rfloor$ 
    % \hfill \COMMENT{Number of retained tokens}
    \STATE $\mathcal{I} \leftarrow \text{TopKIndices}(\mathbf{I}, n)$ \hfill \COMMENT{Indices of pivotal tokens}

    \STATE /* Update cache with selected tokens */
    \STATE $\mathbf{K}_{c} \leftarrow \text{Select}(\mathbf{K}_{f}, \mathcal{I})$
    \STATE $\mathbf{V}_{c} \leftarrow \text{Select}(\mathbf{V}_{f}, \mathcal{I})$
    \STATE $\text{Cache}_l \leftarrow \{ \text{``k"}: \mathbf{K}_{c}, \text{``v"}: \mathbf{V}_{c} \}$
\end{algorithmic}
\end{algorithm}

\subsection{Cache Management for Sparse-dLLM}

To implement the delayed cache updates strategy, we assign a cache state to each step within the current block, where the value can be 0, 1, or 2. $\text{cache\_state} = 0$ indicates performing bidirectional attention for the full sequence at the current step, $\text{cache\_state} = 1$ indicates updating the KV cache, while $\text{cache\_state} = 2$ indicates reusing the KV cache for attention computation. The process of assigning $\text{cache\_state}$ to the decoding step is described in Algorithm \ref{alg:get_cache_state}, and the procedure for cache management based on $\text{cache\_state}$ is presented in Algorithm \ref{alg:cache_management}.

\begin{algorithm}[!ht]
\caption{Assign Cache State for the Current Step}
\label{alg:get_cache_state}
\begin{algorithmic}[1]
    \REQUIRE The decoding step $i$ within the current block.
    \ENSURE The cache state for the current step.
    
    \IF{$i > 1$}
        \STATE $\text{cache\_state} \leftarrow 2$
    \ELSE
        \STATE $\text{cache\_state} \leftarrow i$
    \ENDIF
    \STATE \textbf{return} $\text{cache\_state}$
\end{algorithmic}
\end{algorithm}

\begin{algorithm}[!ht]
\caption{Cache Management Logic}
\label{alg:cache_management}
\begin{algorithmic}[1]
    \REQUIRE Layer id $l$, KV states of the current block $\mathbf{K}_{n}$ and $\mathbf{V}_{n}$, the cache state of the current step $\text{cache\_state}$.
    \ENSURE Cache management for layer $l$.

    \IF{cache\_state == 0}
        \STATE Do nothing \hfill \COMMENT{Cache is not used}
    \ELSIF{cache\_state == 1}
        \STATE Update Cache \hfill \COMMENT{Execute Algorithm~\ref{alg:dynamic_cache_eviction}}
    \ELSIF{cache\_state == 2}
        \STATE $\mathbf{K}_{c} \leftarrow \text{get\_cache}(l)[\text{``k"}]$ \hfill \COMMENT{Reuse cache}
        \STATE $\mathbf{V}_{c} \leftarrow \text{get\_cache}(l)[\text{``v"}]$
        \STATE $\mathbf{K} \leftarrow \text{Concat}(\mathbf{K}_{c}, \mathbf{K}_{n})$
        \STATE $\mathbf{V} \leftarrow \text{Concat}(\mathbf{V}_{c}, \mathbf{V}_{n})$ 
    \ENDIF
\end{algorithmic}
\end{algorithm}

\begin{algorithm}[!t]
\caption{N-Step Delayed Cache Updates}
\label{alg:get_cache_state_for_n_delay}
\begin{algorithmic}[1]
    \REQUIRE The decoding step $i$ within the current block, delay step $x$.
    \ENSURE The cache state for the current step.
    
    \IF{$i > x$}
        \STATE $\text{cache\_state} \leftarrow 2$
    \ELSIF{$i == x$}
        \STATE $\text{cache\_state} \leftarrow 1$
    \ELSE
        \STATE $\text{cache\_state} \leftarrow 0$
    \ENDIF
    \STATE \textbf{return} $\text{cache\_state}$
\end{algorithmic}
\end{algorithm}

\begin{algorithm}[!ht]
\caption{Prefix-Sparse}
\label{alg:prefix-sparse}
\begin{algorithmic}[1]
    \REQUIRE Layer id $l$, query states of the current block $\mathbf{Q}_{b}$, KV states $\mathbf{K}$ and $\mathbf{V}$ for the full sequence, offset of the current block $o$,  block length $b$, retention ratio $r$, kernel size $k$.
    \ENSURE Updated cache for layer $l$. 
    
    \STATE /* Extract KV states before the current block */
    \STATE $\mathbf{K}_{f} \leftarrow \mathbf{K}_{:o}$
    \STATE $\mathbf{V}_{f} \leftarrow \mathbf{V}_{:o}$

    \STATE /* Extract KV states after the current block */
    \STATE $\mathbf{K}_{r} \leftarrow \mathbf{K}_{o+b:}$
    \STATE $\mathbf{V}_{r} \leftarrow \mathbf{V}_{o+b:}$

    \STATE /* Calculate attention scores */
    \STATE $\bar{q} \leftarrow \text{Mean}(\mathbf{Q}_{b})$ \hfill \COMMENT{Average query state in the block}
    \STATE $\mathbf{A} \leftarrow \text{MatMul}(\bar{q}, \mathbf{K}_{f}^T)$ \hfill \COMMENT{Calculate attention scores}

    \STATE /* Max pooling and select tokens */
    \STATE \parbox[t]{\linewidth}{
        $\mathbf{I} \leftarrow \text{MaxPool1D}(\mathbf{A}, \text{kernel\_size}=k, \text{padding}=k/2)$ \hfill \COMMENT{Calculate importance scores}
    }
    \STATE $n \leftarrow \lfloor |\mathbf{I}| \cdot r \rfloor$ 
    % \hfill \COMMENT{Number of retained tokens}
    \STATE $\mathcal{I} \leftarrow \text{TopKIndices}(\mathbf{I}, n)$ \hfill \COMMENT{Indices of pivotal tokens}

    \STATE /* Update cache with selected tokens */
    \STATE $\mathbf{K}_{c} \leftarrow \text{Concat}(\text{Select}(\mathbf{K}_{f}, \mathcal{I}), \mathbf{K}_{r})$
    \STATE $\mathbf{V}_{c} \leftarrow \text{Concat}(\text{Select}(\mathbf{V}_{f}, \mathcal{I}), \mathbf{V}_{r})$
    \STATE $\text{Cache}_l \leftarrow \{ \text{``k"}: \mathbf{K}_{c}, \text{``v"}: \mathbf{V}_{c} \}$
\end{algorithmic}
\end{algorithm}

\subsection{Cache Management for N-Step Delayed Cache Updates}

When modifying the delay step, the process of assigning $\text{cache\_state}$ must be adjusted accordingly, as detailed in Algorithm~\ref{alg:get_cache_state_for_n_delay}.

\subsection{Cache Implementation for Prefix-Sparse}

The prefix-sparse strategy requires: (1) only evicting KV states before the current block, and (2) concatenating complete KV states after the current block when storing KV cache. See Algorithm~\ref{alg:prefix-sparse} for implementation details.

\section{Experiment Details}

\subsection{Benchmarks and Settings}

Table~\ref{tab:benchmark_config} presents the detailed configurations for each benchmark, including the number of decoding steps, block length, and generation length. The benchmarks include MMLU (5-shot), ARC-C (0-shot), PIQA (0-shot), GPQA (5-shot), GSM8K (4-shot), Math (4-shot), and HumanEval (0-shot). To test the generalization and robustness of different approaches, we minimize task-specific hyperparameter tuning and instead adopt a consistent configuration for all benchmarks except HumanEval. Due to its distinct task nature, HumanEval requires a larger number of decoding steps and a longer generation length. 

\begin{table}[ht]
\centering
\begin{tabular}{lccc}
\hline
\textbf{Datasets} & \textbf{Steps} & \textbf{Block Len} & \textbf{Gen Len} \\
\hline
MMLU & 256 & 32 & 256 \\
ARC-C & 256 & 32 & 256 \\
PIQA & 256 & 32 & 256 \\
GPQA & 256 & 32 & 256 \\
GSM8K & 256 & 32 & 256 \\
Math & 256 & 32 & 256 \\
HumanEval & 512 & 32 & 512 \\
\hline
\end{tabular}
\caption{Configuration of Benchmarks}
\label{tab:benchmark_config}  
\end{table}

\subsection{Implementation Details}

In this section, we provide a detailed description of the parameter configurations for the comparative methods dKV-Cache and dLLM-Cache across different models. Following the recommended settings in the dKV-Cache paper, we set the cache refresh interval to 8 for the LLaDA series and to 4 for the Dream series. 

For dLLM-Cache, the paper presents multiple parameter configurations, where $K_p$ denotes the prompt refresh interval and $K_r$ represents the response refresh interval. After comprehensive consideration, we configure the parameters as follows:
\begin{itemize}
    \item For LLaDA-8B-Instruct: $K_p = 50$, $K_r = 7$
    \item For LLaDA-1.5: $K_p = 100$, $K_r = 6$
    \item For Dream-v0-7B-Base: $K_p = 100$, $K_r = 2$
    \item For Dream-v0-7B-Instruct: $K_p = 50$, $K_r = 2$
\end{itemize}

\section{More Results}

\subsection{Long-Context Performance}

To evaluate the performance of different methods on long-context, we conducted experiments using the LongBench benchmark. With the input length truncated to 4k tokens, a block length of 32, and both decoding steps and generation length set to 512, the results are presented in Table~\ref{tab_longbench}. The results demonstrate that Sparse-dLLM has an almost negligible impact on the model's long-context capability.

\begin{table*}[ht]
\centering
\begin{tabular}{lccccc}
\toprule
\textbf{} & Base & +dLLM-Cache & +dKV-Cache & +Fast-dLLM & +Sparse-dLLM \\
\midrule
- \textbf{LLaDA-8B-Instruct} & 34.55 & 33.34 & 34.72 & 33.99 & 34.46 \\
- \textbf{LLaDA-1.5} & 34.66 & 33.65 & 34.68 & 34.33 & 34.50 \\
\midrule
- \textbf{Dream-v0-7B-Base} & 34.17 & 34.11 & 34.16 & 33.54 & 34.13 \\
- \textbf{Dream-v0-7B-Instruct} & 38.62 & 38.49 & 38.56 & 38.00 & 38.30 \\
\bottomrule
\end{tabular}
\caption{Experimental Results on LongBench.}
\label{tab_longbench}
\end{table*}

\section{More Ablations and Analysis}

\subsection{Hyperparameters: Retention Ratio and Kernel Size}

Here are the experimental results evaluating model performance on GSM8K (4-shot) across various retention ratios ($r$) and kernel sizes ($k$). Table~\ref{tab_gsm8k_diff_r_k} shows the accuracy of LLaDA-8B-Instruct under different configurations.

\begin{table*}[!htb]
\centering
\begin{tabular}{lccccccccc}
\toprule
 & $r=0.1$ & $r=0.2$ & $r=0.3$ & $r=0.4$ & $r=0.5$ & $r=0.6$ & $r=0.7$ & $r=0.8$ & $r=0.9$ \\ 
\midrule
$k=1$ & \textbf{73.69} & \textbf{76.80} & \textbf{77.03} & \textbf{77.10} & \underline{77.26} & 77.33 & 76.42 & \textbf{78.01} & 77.18 \\
$k=3$ & \underline{67.02} & 75.89 & 75.59 & 76.35 & \textbf{77.56} & 76.95 & 76.95 & 76.35 & \textbf{77.56} \\
$k=5$ & 52.77 & 75.97 & 76.27 & \underline{76.57} & 76.42 & \underline{77.41} & \textbf{77.10} & \underline{77.33} & 77.10 \\
$k=7$ & 42.76 & \underline{76.04} & \underline{76.42} & 76.27 & 76.95 & \textbf{77.63} & 76.65 & 76.65 & \underline{77.48} \\
$k=9$ & 35.48 & 75.44 & 76.35 & 75.51 & \underline{77.26} & 77.10 & \underline{77.03} & 76.27 & 76.42 \\
\bottomrule
\end{tabular}
\caption{GSM8K (4-shot) accuracy (\%) on LLaDA-8B-Instruct across retention ratios ($r$) and kernel sizes ($k$). Bold/underline denotes top-1/top-2 performance per $r$.}
\label{tab_gsm8k_diff_r_k}
\end{table*}

\subsection{Analysis of Pivotal tokens}

\begin{figure*}[!h]
\centering
\begin{subfigure}[b]{0.48\linewidth}
    \centering
    \includegraphics[width=0.92\linewidth]{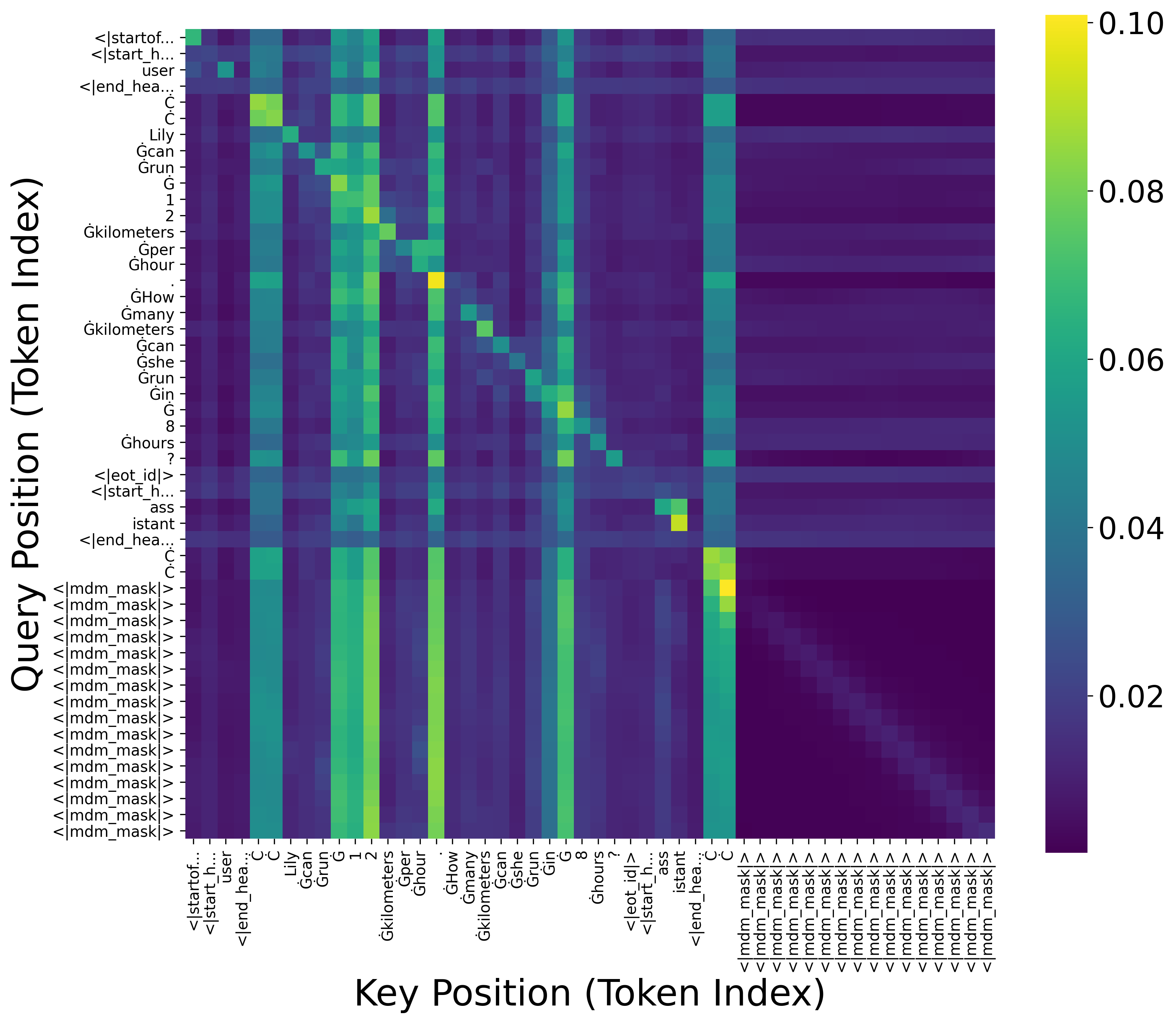}
    \caption{Layer 0, Step 0}
    \label{fig_pivotal_tokens_0_0}
\end{subfigure}
\hfill
\begin{subfigure}[b]{0.48\linewidth}
    \centering
    \includegraphics[width=0.92\linewidth]{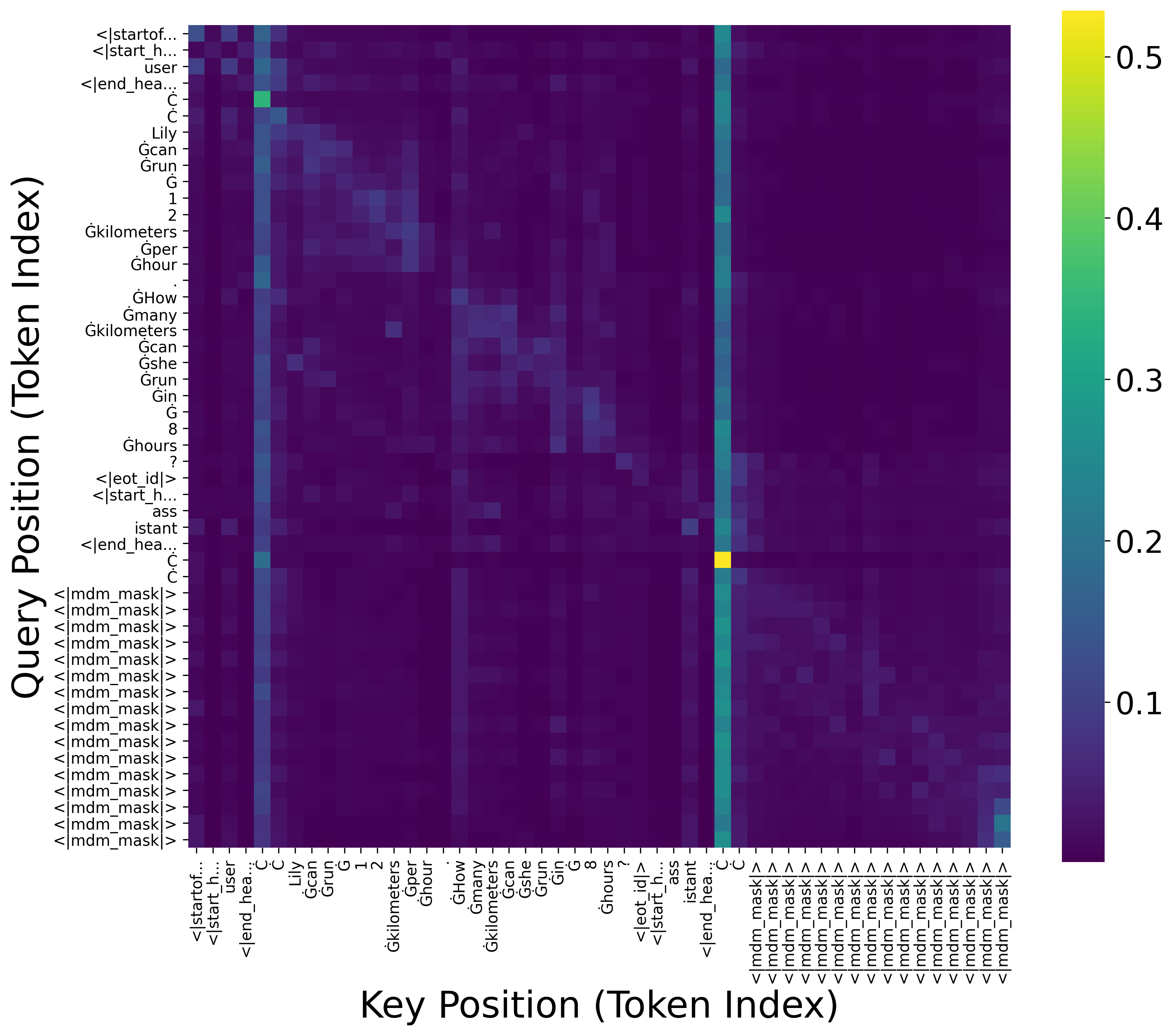}
    \caption{Layer 15, Step 0}
    \label{fig_pivotal_tokens_31_0}
\end{subfigure}
\caption{Attention heatmaps with corresponding labels.}
\label{fig_pivotal_tokens}
\end{figure*}

To investigate the distribution of pivotal tokens, we further examined the attention heatmaps annotated with labels, as shown in Figure~\ref{fig_pivotal_tokens}. The results indicate that the majority of pivotal tokens are those carrying no semantic information, such as line breaks and spaces. This observation may inspire future research on more fine-grained acceleration approaches for dLLMs.

\subsection{Local Aggregation: AvgPool vs. MaxPool}

To investigate the impact of different pooling operations on model performance, we conducted evaluations on the LLaDA-8B-Instruct model, as shown in Table~\ref{tab_avgpool_vs_maxpool}. The experimental results demonstrate marginal differences between the two pooling operations, with both leading to improved model performance.

\begin{table}[!h]
\centering
\tabcolsep=0.18cm
\small
\begin{tabular}{lcccccccc}
\toprule
 & MMLU & ARC-C & PIQA & GPQA & GSM8k & Math & HE & Avg. \\
\midrule 
\textbf{LLaDA-8B-Instruct} & 60.60 & \underline{88.47} & 83.62 & 32.83 & \textbf{78.39} & \textbf{36.02} & 34.76 & 59.24 \\
+Sparse-dLLM w/ AvgPool & \textbf{61.20} & \textbf{88.81} & \textbf{84.60} & \textbf{35.86} & 77.48 & \underline{34.50} & \underline{35.98} & \underline{59.78} \\
+Sparse-dLLM w/ MaxPool & \underline{61.01} & \underline{88.47} & \underline{84.44} & \underline{35.35} & \underline{77.56} & 34.42 & \textbf{37.80} & \textbf{59.86} \\
\bottomrule
\end{tabular}
\caption{Experimental results on different pooling operations. Best values in bold,
suboptimal values underlined.}
\label{tab_avgpool_vs_maxpool}
\end{table}

\end{document}